\pdfoutput=1

\PassOptionsToPackage{unicode}{hyperref}
\PassOptionsToPackage{hyphens}{url}
\documentclass[
]{article}
\usepackage{amsmath,amssymb}
\usepackage{lmodern}
\usepackage[T1]{fontenc}
\usepackage[utf8]{inputenc}
\usepackage{textcomp}
\usepackage{lmodern} 
\IfFileExists{upquote.sty}{\usepackage{upquote}}{}
\IfFileExists{microtype.sty}{% use microtype if available
  \usepackage[]{microtype}
  \UseMicrotypeSet[protrusion]{basicmath} % disable protrusion for tt fonts
}{}
\makeatletter
\@ifundefined{KOMAClassName}{% if non-KOMA class
  \IfFileExists{parskip.sty}{%
    \usepackage{parskip}
  }{% else
    \setlength{\parindent}{0pt}
    \setlength{\parskip}{6pt plus 2pt minus 1pt}}
}{% if KOMA class
  \KOMAoptions{parskip=half}}
\makeatother
\usepackage{xcolor}
\usepackage{longtable,booktabs,array}
\usepackage{calc} % for calculating minipage widths
\usepackage{etoolbox}
\makeatletter
\patchcmd\longtable{\par}{\if@noskipsec\mbox{}\fi\par}{}{}
\makeatother
\IfFileExists{footnotehyper.sty}{\usepackage{footnotehyper}}{\usepackage{footnote}}
\makesavenoteenv{longtable}
\providecommand{\tightlist}{%
  \setlength{\itemsep}{0pt}\setlength{\parskip}{0pt}}
\IfFileExists{bookmark.sty}{\usepackage{bookmark}}{\usepackage{hyperref}}
\IfFileExists{xurl.sty}{\usepackage{xurl}}{} % add URL line breaks if available
\hypersetup{
  hidelinks,
  pdfcreator={LaTeX via pandoc}}

\title{World Consistency Score: A Unified Metric for Video Generation Quality}

\author{
  Akshat Rakheja\thanks{Equal contribution} \and 
  Aarsh Ashdhir\footnotemark[1] \and 
  Aryan Bhattacharjee \and 
  Vanshika Sharma
}

\date{July 31, 2025}

\begin{document}

\maketitle

\hypertarget{abstract}{%
\subsection{Abstract}\label{abstract}}

We introduce \textbf{World Consistency Score (WCS)}, a novel unified
evaluation metric for generative video models that emphasizes
\emph{internal world consistency} of the generated videos. WCS
integrates four interpretable sub-components -- \textbf{object
permanence}, \textbf{relation stability}, \textbf{causal compliance},
and \textbf{flicker penalty} -- each measuring a distinct aspect of
temporal and physical coherence in a video. These submetrics are
combined via a learned weighted formula to produce a single consistency
score that aligns with human judgments. We detail the motivation for WCS
in the context of existing video evaluation metrics, formalize each
submetric and how it is computed with open-source tools (trackers,
action recognizers, CLIP embeddings, optical flow), and describe how the
weights of the WCS combination are trained using human preference data.
We also outline an experimental validation blueprint: using benchmarks
like VBench-2.0, EvalCrafter, and LOVE to test WCS's correlation with
human evaluations, performing sensitivity analyses, and comparing WCS
against established metrics (FVD, CLIPScore, VBench, FVMD). The proposed
WCS offers a comprehensive and interpretable framework for evaluating
video generation models on their ability to maintain a coherent
``world'' over time, addressing gaps left by prior metrics focused only
on visual fidelity or prompt alignment.

\hypertarget{introduction}{%
\subsection{1. Introduction}\label{introduction}}

Generative video models have seen rapid progress with the advent of
large-scale diffusion models and multimodal transformers, achieving
impressive visual quality and even basic prompt-following
ability\href{https://arxiv.org/html/2503.21765v1\#:~:text=44\%20\%2C\%20\%2080\%2C\%2046,success\%20in\%20many\%20downstream\%20tasks}{{[}1{]}}\href{https://www.reddit.com/r/artificial/comments/1jmgy6n/vbench20_a_framework_for_evaluating_intrinsic/\#:~:text=I\%20think\%20this\%20work\%20represents,quality\%20metrics\%20alone\%20would\%20indicate}{{[}2{]}}.
However, a persistent shortcoming is that \emph{``generated content
often violates the fundamental laws of physics, falling into the dilemma
of `visual realism but physical
absurdity'\,''}\href{https://arxiv.org/html/2503.21765v1\#:~:text=especially\%20with\%20the\%20rapid\%20advancement,a\%20comprehensive\%20summary\%20of\%20architecture}{{[}3{]}}.
Videos generated by state-of-the-art models may look sharp and
temporally smooth, yet upon closer inspection, they suffer from
\textbf{world inconsistency} -- objects appear or disappear arbitrarily,
spatial relationships change without cause, and physical causality is
not respected. These issues lead to videos that a human viewer would
judge as incoherent or implausible, even if individual frames are
photorealistic.

Existing evaluation metrics do not adequately capture these consistency
issues. The widely used \emph{Fréchet Video Distance (FVD)} compares
distributions of deep features between generated and real
videos\href{https://qiyan98.github.io/blog/2024/fvmd-1/\#:~:text=Fr\%C3\%A9chet\%20Video\%20Distance\%20,Nevertheless\%2C\%20both\%20are}{{[}4{]}},
which is useful for overall quality assessment but fails to pinpoint
logical errors within a single video. \emph{CLIPScore} and related
measures evaluate semantic alignment (e.g., between video frames and a
text
prompt)\href{https://qiyan98.github.io/blog/2024/fvmd-1/\#:~:text=For\%20text,the\%20average\%20similarity\%20between\%20each}{{[}5{]}},
but do not directly assess temporal coherence or physical realism.
Recently, benchmark suites like \emph{VBench} and \emph{EvalCrafter}
have begun to incorporate finer-grained evaluation dimensions -- e.g.,
identity consistency, motion smoothness, spatial relations,
etc.\href{https://vchitect.github.io/VBench-project/\#:~:text=quality,3}{{[}6{]}}\href{https://evalcrafter.github.io/\#:~:text=generation\%20by\%20analyzing\%20the\%20real,method\%2C\%20our\%20method\%20can\%20successfully}{{[}7{]}}
-- and use weighted combinations of metrics validated against human
preferences\href{https://qiyan98.github.io/blog/2024/fvmd-1/\#:~:text=VBench\%20proposes\%20a\%20comprehensive\%20set,the\%20effectiveness\%20of\%20these\%20metrics}{{[}8{]}}.
VBench-2.0, for example, evaluates text-to-video fidelity across seven
aspects (object presence, actions, spatial/temporal relations, etc.) and
showed strong correlation (Pearson \textgreater{} 0.7) between their
automatic metrics and human
judgments\href{https://www.reddit.com/r/artificial/comments/1jmgy6n/vbench20_a_framework_for_evaluating_intrinsic/\#:~:text=,prompt\%20templates\%2C\%20generating\%205\%2C700\%2B\%20videos}{{[}9{]}}.
These efforts confirm that a multi-dimensional approach can better align
with human perception than any single metric.

\textbf{Our Contribution:} In this work, we propose the \textbf{World
Consistency Score (WCS)}, a unified no-reference metric designed to
capture the \emph{internal consistency and physical plausibility} of a
generated video. Rather than requiring a ground-truth video or focusing
only on prompt compliance, WCS analyzes the generated video as a
self-contained ``world'' and evaluates:

\begin{itemize}
\tightlist
\item
  \textbf{Object Permanence:} Do objects that appear in the video
  continue to exist consistently, even when briefly out of sight? (No
  inexplicable vanishing or appearing of objects.)
\item
  \textbf{Relation Stability:} Are spatial and structural relationships
  between entities maintained over time, absent deliberate changes? (No
  sudden jumps in relative positions or sizes without cause.)
\item
  \textbf{Causal Compliance:} Do events in the video obey
  cause-and-effect constraints from physics and common sense? (No
  effects happening without causes, and no causes without expected
  effects.)
\item
  \textbf{Flicker Penalty:} Is the video free of unnatural flicker or
  jitter? (No rapid, unexplained pixel changes frame-to-frame that
  disrupt temporal smoothness.)
\end{itemize}

Each of these components addresses a key aspect of what makes a video's
content \emph{consistent}. Together, they form a comprehensive view of a
video model's \emph{world modeling} capability -- i.e., whether the
model has preserved an intelligible and stable ``world'' in the sequence
it
generates\href{https://www.aimodels.fyi/papers/arxiv/worldmodelbench-judging-video-generation-models-as-world\#:~:text=Today\%27s\%20AI\%20systems\%20that\%20generate,systems\%20actually\%20understand\%20physical\%20reality}{{[}10{]}}\href{https://www.aimodels.fyi/papers/arxiv/worldmodelbench-judging-video-generation-models-as-world\#:~:text=in\%20these\%20scenes\%3F}{{[}11{]}}.
Crucially, we formulate WCS as a \textbf{learned weighted combination}
of these submetrics, allowing it to be tuned to human subjective
preferences by regression on human ratings data. This approach draws
inspiration from recent learned evaluators like PickScore (for images)
and human-aligned video
benchmarks\href{https://evalcrafter.github.io/\#:~:text=generation\%20by\%20analyzing\%20the\%20real,method\%2C\%20our\%20method\%20can\%20successfully}{{[}7{]}}\href{https://qiyan98.github.io/blog/2024/fvmd-1/\#:~:text=is\%20determined\%20by\%20a\%20weighted,the\%20effectiveness\%20of\%20these\%20metrics}{{[}12{]}},
aiming to maximize correlation with what people consider a coherent,
high-quality video.

In the following, we first review related metrics and motivate the need
for WCS (Section 2). We then introduce the formal definitions of the
four submetrics (Section 3), detailing how each can be computed using
open-source tools (object trackers, action recognizers, CLIP encoders,
and optical flow). We present the WCS formula, explaining how the
weights are learned from human preference data (Section 4). Next, we
outline an experimental design for validating WCS (Section 5), including
benchmark datasets (VBench-2.0, EvalCrafter, LOVE), planned sensitivity
analyses, ablations, and comparisons to existing metrics like FVD,
CLIPScore, VBench and FVMD. Finally, we discuss the implications of WCS
and conclude with future directions (Section 6). Our aim is to provide a
self-contained, rigorous yet accessible description of this new metric
-- focusing on \emph{motivation, methodology, and `evaluation blueprint}
-- in lieu of extensive quantitative results (which we defer to future
work once the metric is implemented).

By introducing WCS, we hope to push video generation evaluation beyond
surface-level quality towards deeper semantic and physical consistency.
A high WCS would indicate a video generation that not only looks good
frame-by-frame, but also ``makes sense'' as a depiction of a plausible
world over time. We believe this is a critical step toward generative
models with true understanding of temporal and physical dynamics, rather
than just producing momentarily pretty
pictures\href{https://www.aimodels.fyi/papers/arxiv/worldmodelbench-judging-video-generation-models-as-world\#:~:text=struggled\%20with\%20long,track\%20it\%20over\%20longer\%20periods}{{[}13{]}}\href{https://www.aimodels.fyi/papers/arxiv/worldmodelbench-judging-video-generation-models-as-world\#:~:text=The\%20research\%20also\%20found\%20that,the\%20underlying\%20physics\%20and\%20dynamics}{{[}14{]}}.

\hypertarget{background-and-related-work}{%
\subsection{2. Background and Related
Work}\label{background-and-related-work}}

\textbf{Conventional Video Quality Metrics:} Early metrics for video
generation borrowed from video compression and processing literature,
focusing on low-level fidelity. Examples include \emph{PSNR} and
\emph{SSIM}, which compare pixel-level differences or structural
similarity to a reference
video\href{https://qiyan98.github.io/blog/2024/fvmd-1/\#:~:text=In\%20addition\%20to\%20these\%20modern,trained\%20models}{{[}15{]}}.
However, such metrics fail to account for generative realism or temporal
coherence. More advanced reference-based metrics like \emph{Fréchet
Inception Distance (FID)} and its video extension \emph{Fréchet Video
Distance (FVD)} are distribution-oriented: they embed videos (or frames)
into a feature space (often using pretrained CNN or 3D ConvNet features)
and compute a statistical distance (Fréchet) between generated samples
and real
samples\href{https://qiyan98.github.io/blog/2024/fvmd-1/\#:~:text=Fr\%C3\%A9chet\%20Inception\%20Distance\%20,the\%20temporal\%20coherence\%20between\%20frames}{{[}16{]}}\href{https://qiyan98.github.io/blog/2024/fvmd-1/\#:~:text=Fr\%C3\%A9chet\%20Video\%20Distance\%20,Nevertheless\%2C\%20both\%20are}{{[}4{]}}.
FVD's use of an I3D 3D ConvNet (trained on action recognition) makes it
somewhat sensitive to temporal aberrations (e.g. frame scrambling) in
addition to overall visual
quality\href{https://qiyan98.github.io/blog/2024/fvmd-1/\#:~:text=Fr\%C3\%A9chet\%20Video\%20Distance\%20,Nevertheless\%2C\%20both\%20are}{{[}4{]}}.
Indeed, FVD was found to align better with human judgment than a purely
spatial FID or than the alternative Kernel Video
Distance\href{https://qiyan98.github.io/blog/2024/fvmd-1/\#:~:text=but\%20also\%20to\%20temporal\%20aberrations,metrics\%20for\%20unconditional\%20video\%20generation}{{[}17{]}}.
Nevertheless, FVD is a \emph{set-level} metric: it evaluates a model's
output distribution against ground-truth distribution \emph{across many
videos}, and does not offer a meaningful assessment of a single video's
logical consistency. A model could score well on FVD by averaging out
errors across samples, or by producing videos that are statistically
plausible but internally inconsistent in ways that do not drastically
affect the feature distribution.

\textbf{Semantic Similarity Metrics:} With the rise of text-to-video
generation, evaluating \emph{text-video alignment} became important.
\emph{CLIPScore} is an example unary metric using a pretrained
multimodal model (CLIP) to measure how well video frames match the
textual
prompt\href{https://qiyan98.github.io/blog/2024/fvmd-1/\#:~:text=For\%20text,the\%20average\%20similarity\%20between\%20each}{{[}5{]}}.
Typically, one computes the cosine similarity between CLIP embedding of
each frame and the prompt, averaging over the
video\href{https://qiyan98.github.io/blog/2024/fvmd-1/\#:~:text=the\%20embedding\%20space,frame\%20accuracy\%29}{{[}18{]}}.
A high CLIPScore indicates the video content is relevant to the prompt
(e.g. if prompt says ``a cat on a sofa'', the frames indeed contain such
scene). However, CLIPScore does not penalize temporal inconsistencies
\emph{unless} they cause a change in the overall semantic content. Some
works also use CLIP to assess temporal coherence by measuring similarity
between adjacent frames'
embeddings\href{https://qiyan98.github.io/blog/2024/fvmd-1/\#:~:text=video\%20sequence\%20adheres\%20to\%20the,frame\%20accuracy\%29}{{[}19{]}}
-- if consecutive frames have very different CLIP image embeddings, that
might indicate a flicker or a scene change. This ``CLIP frame
consistency'' idea is useful, but CLIP embeddings are coarse and not
explicitly designed to catch physical errors; e.g., a scene where a cat
teleports across the room might still yield similar frame embeddings
(both frames contain ``a cat in a room''). Thus, semantic metrics alone
are insufficient for detailed world consistency evaluation.

\textbf{Comprehensive Benchmark Suites:} Recent efforts have moved
toward \emph{multi-faceted evaluation}.
\textbf{VBench}\href{https://vchitect.github.io/VBench-project/\#:~:text=Video\%20generation\%20has\%20witnessed\%20significant,inconsistency\%2C\%20motion\%20smoothness\%2C\%20temporal\%20flickering}{{[}20{]}}
introduced \emph{16 dimensions} of video quality, covering aspects like
\emph{subject identity consistency, object fidelity, motion stability,
temporal flicker, spatial relations,} and
more\href{https://vchitect.github.io/VBench-project/\#:~:text=quality,3}{{[}6{]}}.
Each dimension is evaluated with a specialized method or model (for
example, using an optical flow model to measure motion smoothness, using
detection models to count objects, etc.), and an overall score can be
formed by weighting these
dimensions\href{https://qiyan98.github.io/blog/2024/fvmd-1/\#:~:text=VBench\%20proposes\%20a\%20comprehensive\%20set,the\%20effectiveness\%20of\%20these\%20metrics}{{[}8{]}}.
Notably, VBench provided human preference annotations for each dimension
and showed that their automated metrics correlate well with human
perception on those
axes\href{https://vchitect.github.io/VBench-project/\#:~:text=and\%20spatial\%20relationship\%2C\%20etc\%29,quality\%20Image\%20Suite\%20with\%20an}{{[}21{]}}.
The follow-up \textbf{VBench-2.0} focused on \emph{intrinsic
faithfulness} to text
prompts\href{https://www.reddit.com/r/artificial/comments/1jmgy6n/vbench20_a_framework_for_evaluating_intrinsic/\#:~:text=VBench,generation\%20models\%20against\%20these\%20metrics}{{[}22{]}},
defining seven metrics such as object presence, attribute correctness,
action accuracy, \emph{spatial relation} and \emph{temporal relation}
faithfulness,
etc.\href{https://www.reddit.com/r/artificial/comments/1jmgy6n/vbench20_a_framework_for_evaluating_intrinsic/\#:~:text=,Temporal\%20Relation\%2C\%20and\%20Background\%20Faithfulness}{{[}23{]}}.
While prompt-focused, these include checking if the relative spatial
configuration and sequence of events in the video match the described
relations -- conceptually similar to our relation stability and causal
compliance but always referenced to an explicit prompt.
\textbf{EvalCrafter}\href{https://evalcrafter.github.io/\#:~:text=For\%20video\%20generation\%2C\%20various\%20open,the\%20help\%20of\%20the\%20large}{{[}24{]}}
likewise introduced a pipeline evaluating videos on \emph{visual
quality, content quality, motion quality, temporal consistency,} and
\emph{text alignment} with \textbf{17 objective metrics}, then performed
\textbf{subjective studies} for each aspect and learned a mapping from
the objective metrics to a final \emph{``user favor''}
score\href{https://evalcrafter.github.io/\#:~:text=generation\%20by\%20analyzing\%20the\%20real,method\%2C\%20our\%20method\%20can\%20successfully}{{[}7{]}}.
Their approach of aligning metrics to human opinions via regression is
directly relevant to how we learn WCS weights. Other benchmarks like
\textbf{LOVE} (AIGV-Eval dataset) aggregate human Mean Opinion Scores
(MOS) on dimensions like perceptual quality and
correspondence\href{https://www.researchgate.net/publication/391878222_LOVE_Benchmarking_and_Evaluating_Text-to-Video_Generation_and_Video-to-Text_Interpretation\#:~:text=present\%20AIGVE,Comprehensive\%20experiments\%20demonstrate\%20that}{{[}25{]}},
and train large multimodal evaluators to predict these
scores\href{https://www.researchgate.net/publication/391878222_LOVE_Benchmarking_and_Evaluating_Text-to-Video_Generation_and_Video-to-Text_Interpretation\#:~:text=V2T\%20interpretation\%20capabilities,codes\%20are\%20anonymously\%20available\%20at}{{[}26{]}}.
Those learned evaluators (e.g. the LOVE metric) treat the problem as a
multi-task learning using a single model (like a video-language model)
to output
scores\href{https://www.researchgate.net/publication/391878222_LOVE_Benchmarking_and_Evaluating_Text-to-Video_Generation_and_Video-to-Text_Interpretation\#:~:text=V2T\%20interpretation\%20capabilities,com\%2FIntMeGroup\%2FLOVE}{{[}27{]}}.
In contrast, WCS embraces a \emph{modular}, interpretable design: we
hand-craft submetrics capturing physics and consistency, which could
then be linearly combined. This makes WCS more explainable -- one can
inspect which aspect of consistency a model fails at -- whereas
end-to-end learned metrics (LOVE, etc.) act as black boxes (albeit
powerful ones).

\textbf{Physical Consistency and World Models:} A few recent works
explicitly address physical realism in generated videos. \emph{Fréchet
Video Motion Distance (FVMD)} was proposed to focus specifically on
motion
dynamics\href{https://github.com/DSL-Lab/FVMD-frechet-video-motion-distance\#:~:text=We\%20propose\%20the\%20Fr\%C3\%A9chet\%20video,motion\%20consistency\%20of\%20video\%20generation}{{[}28{]}}.
FVMD tracks keypoints in videos and compares the distributions of
velocities and accelerations between generated and real
videos\href{https://qiyan98.github.io/blog/2024/fvmd-1/\#:~:text=Fr\%C3\%A9chet\%20Video\%20Motion\%20Distance\%20,laws\%20and\%20avoid\%20abrupt\%20changes}{{[}29{]}}.
By measuring whether motion is smooth and follows typical physics (no
abrupt jerks, accelerations within reasonable ranges), FVMD targets
\emph{motion
consistency}\href{https://qiyan98.github.io/blog/2024/fvmd-1/\#:~:text=Fr\%C3\%A9chet\%20Video\%20Motion\%20Distance\%20,laws\%20and\%20avoid\%20abrupt\%20changes}{{[}29{]}}.
This aligns with part of our \emph{causal compliance} and \emph{flicker}
aims -- however, FVMD still needs a set of real videos for comparison,
and it reduces the rich notion of ``physical plausibility'' to a
statistical similarity of motion patterns. In our view, a model could
theoretically cheat FVMD by producing \emph{consistently} wrong physics
(e.g. always slightly low gravity), and FVMD might not flag it if it's
statistically consistent. We prefer a more direct evaluation of
causality (checking cause-effect within a single video). Meanwhile,
\emph{WorldModelBench}\href{https://www.aimodels.fyi/papers/arxiv/worldmodelbench-judging-video-generation-models-as-world\#:~:text=,term}{{[}30{]}}
treated video generative models as world simulators and evaluated
capabilities like physical reasoning, object permanence, and action
understanding by having models predict future frames and comparing to
ground truth
videos\href{https://www.aimodels.fyi/papers/arxiv/worldmodelbench-judging-video-generation-models-as-world\#:~:text=WorldModelBench\%20evaluates\%20models\%20by\%20having,accidents\%2C\%20and\%20kitchen\%20interactions\%20respectively}{{[}31{]}}\href{https://www.aimodels.fyi/papers/arxiv/worldmodelbench-judging-video-generation-models-as-world\#:~:text=1,understanding\%3A\%20Maintaining\%20consistent\%20object\%20representations}{{[}32{]}}.
They introduced a metric called \textbf{Mutual Segment Matching (MSM)}
which compares object segmentation masks between generated and real
sequences to see if objects are accurately tracked and
represented\href{https://www.aimodels.fyi/papers/arxiv/worldmodelbench-judging-video-generation-models-as-world\#:~:text=The\%20researchers\%20developed\%20a\%20novel,level\%20accuracy}{{[}33{]}}.
WorldModelBench found current models struggle especially with
\textbf{object permanence} -- \emph{models often fail to track objects
once they disappear from
view}\href{https://www.aimodels.fyi/papers/arxiv/worldmodelbench-judging-video-generation-models-as-world\#:~:text=The\%20models\%20performed\%20best\%20on,when\%20they\%20can\%27t\%20see\%20them}{{[}34{]}}
-- reinforcing the importance of this aspect. However, their evaluation
is reference-based (requiring ground truth future frames) and geared to
future prediction scenarios. In contrast, WCS is designed for
\emph{unconditional or text-conditional generation} where no ground
truth video is available; we instead rely on internal consistency checks
on the generated video itself.

\textbf{Summary:} There is a clear trend towards richer evaluation
protocols that decompose video quality into meaningful dimensions and
attempt to align with human
perception\href{https://vchitect.github.io/VBench-project/\#:~:text=quality,current\%20models\%27\%20ability\%20across\%20various}{{[}35{]}}\href{https://evalcrafter.github.io/\#:~:text=generation\%20by\%20analyzing\%20the\%20real,method\%2C\%20our\%20method\%20can\%20successfully}{{[}7{]}}.
Our proposed WCS falls squarely in this trend, focusing specifically on
\emph{world consistency dimensions}. It can be seen as distilling some
of these recent ideas (object permanence, no flicker, stable relations,
causal logic) into four concise metrics and then learning an optimal
weighting. By doing so, WCS aims to offer both \emph{interpretability}
(scores for each sub-aspect) and \emph{convenience} (a single unified
score). In the next section, we define each submetric formally.

\hypertarget{components-of-the-world-consistency-score}{%
\subsection{3. Components of the World Consistency
Score}\label{components-of-the-world-consistency-score}}

To evaluate whether a generated video is consistent and plausible, we
break the problem into four \textbf{submetrics}. Each submetric
evaluates a different property of the video along the time axis. We
emphasize that all are \emph{no-reference metrics}: they operate on the
generated video alone, without needing a ground-truth video or
predefined prompt, making WCS broadly applicable. Here we describe and
formalize each component, along with how it can be computed using
available tools.

\hypertarget{object-permanence-op}{%
\subsubsection{3.1 Object Permanence (OP)}\label{object-permanence-op}}

\textbf{Definition:} \emph{Object permanence} measures whether objects
persist over time in the video -- once an object is introduced, it
should continue to exist unless there is a plausible reason for it to
disappear (such as leaving the frame or being occluded by another
object). This concept is rooted in cognitive development: humans
understand that objects don't vanish when out of sight. For video
generation, we want to penalize ``magic'' disappearances or appearances
of objects. A model that forgets to render a character after a few
frames, or that pops new items into a scene without explanation,
violates object permanence.

\textbf{Metric Formalization:} At a high level, we want to quantify how
consistently objects are present throughout the video. Let the video consist of frames $I_{1}, I_{2}, \ldots, I_{T}$. First, we need to detect
and track objects. We can apply an \emph{off-the-shelf object detector
and tracker} to the
video\href{https://www.aimodels.fyi/papers/arxiv/worldmodelbench-judging-video-generation-models-as-world\#:~:text=1,understanding\%3A\%20Maintaining\%20consistent\%20object\%20representations}{{[}32{]}}.
For example, one could use a detection model (like YOLOv8 or DINO) on
the first frame to identify distinct objects, and then use a
multi-object tracking algorithm (e.g. ByteTrack or a SORT tracker with
re-identification) to follow each detected object's trajectory through
subsequent frames. Modern trackers, possibly augmented with segmentation
(like SAM) for precise masks, can handle moderate camera motion and
occlusion.

Let $N$ be the number of unique objects detected in the video
(counting only those visible in at least one frame). For each object
$i$ (with $i = 1, \ldots, N$), suppose it is visible in a subset of frames.
Define the \textit{visibility sequence} $v_i(t)$ for object $i$ as:

\[
v_i(t) =
\begin{cases}
1, & \text{if object } i \text{ is visible in frame } t \text{ (according to the tracker)} \\
0, & \text{otherwise}
\end{cases}
\]

Ideally, for strong object permanence, once $v_i(t)$ becomes 1
(the object appears), it should remain 1 for all subsequent frames
until the object logically exits the scene (e.g., walks out of frame).
Abrupt transitions from 1 to 0 (object vanished) or 0 to 1 (object appeared)
are considered suspect unless they coincide with entering/exiting
frame boundaries or occlusion events.

One simple formulation of object permanence score could be: \emph{the
fraction of frames each object remains present after its first
appearance, averaged over objects}. 

For example, if object $i$ first appears at frame $t_i^{\text{start}}$, and
remains visible through $t_i^{\text{end}}$, then the persistence ratio for object $i$ is:
\[
\frac{t_i^{\text{end}} - t_i^{\text{start}} + 1}{T - t_i^{\text{start}} + 1}
\]

If an object persists till the last frame, this ratio is 1 (perfect). 
If it disappears early (large gap before end), the ratio drops. 
We then define Object Permanence (OP) as:

\[\text{OP}\mspace{6mu} = \mspace{6mu}\frac{1}{N}\sum_{i = 1}^{N}\frac{\sum_{t = t_{i}^{\text{start}}}^{T}v_{i}(t)}{T - t_{i}^{\text{start}} + 1},\]

which is essentially the average fraction of time that objects exist
after being introduced. A score of 1 means every object, once
introduced, stayed present for the rest of the video. Lower scores
indicate some objects disappeared prematurely.

To make this more robust, we should handle cases of intentional
disappearance. If an object \emph{exits the field of view} (e.g., walks
out of the camera frame at the edge) or is \emph{fully occluded by
another object}, those are not penalized (that's a plausible
disappearance). We can attempt to detect such cases: if the tracker
shows the object moving towards a boundary in the last few frames it's
seen, we assume it left the frame naturally; or if another object's mask
covers it entirely, we assume occlusion. These nuances can be encoded by
weighting the disappearance less if it matches an exit/occlusion
scenario. In the absence of explicit occlusion modeling, a simpler
heuristic is to ignore objects that disappear at the last frame (could
be leaving scene) and penalize those that vanish in the middle of the
video with no reappearance.

\textbf{Computation Tools:} Implementing OP is feasible with open tools.
We will use a \emph{pretrained detection model} to find objects in each
frame, then link detections into trajectories. The \textbf{TRACK Any
Object} framework or recent trackers with \emph{object permanence
modeling} can even predict temporary
occlusions\href{https://openaccess.thecvf.com/content/ICCV2021/papers/Tokmakov_Learning_To_Track_With_Object_Permanence_ICCV_2021_paper.pdf\#:~:text=,operate\%20in\%20the}{{[}36{]}}.
For example, a tracker that can handle occlusion might continue a track
through short disappearances, effectively ``hallucinating'' through
occlusion, as in OMC (Object Motion Consistency) trackers. Additionally,
to avoid identity switches (tracker mistaking one object for another),
we can incorporate appearance embeddings -- e.g. CLIP embeddings for
each detected object crop. By comparing an object's crop embedding
across frames (cosine similarity), we ensure the tracker's identity for
that object remains consistent (this helps detect if what appears to be
a ``new'' object is actually the same one re-appearing). Such an
approach, using CLIP features for re-ID, can further boost the
reliability of OP measurement, as it reduces false counting of an object
as disappeared when it was actually present but mislabeled as a
different object.

Ultimately, the OP submetric provides a quantitative gauge of how well a
model maintains the existence of entities over time. It directly
penalizes egregious failures like an object present in frame 1 simply
vanishing in frame 2 with no explanation -- a phenomenon unfortunately
common in early text-to-video generations.

\hypertarget{relation-stability-rs}{%
\subsubsection{3.2 Relation Stability
(RS)}\label{relation-stability-rs}}

\textbf{Definition:} \emph{Relation stability} assesses whether the
\emph{relationships between objects} in the video remain stable and
coherent over time. In a physical world, objects maintain consistent
spatial relationships unless an interaction or movement causes a
deliberate change. For example, if a person is sitting on a chair in one
frame, we expect them to remain on the chair and not float above it or
teleport to its side in the next frame. If two objects are separate,
they shouldn't merge into one or swap positions inexplicably. Relation
stability covers both \emph{spatial relations} (relative positions,
orientations, contacts) and possibly \emph{identity relations} (e.g.
which object is holding which, or a part-whole relationship) as the
video progresses.

\textbf{Metric Formalization:} To quantify relation stability, we rely
on tracking multiple objects and monitoring their pairwise spatial
relationships. Using the same tracking setup as in the Object Permanence (OP) metric,
assume that we have trajectories for all major objects. For any pair of objects
$i$ and $j$ that coexist in frames, we can examine metrics such as:

\begin{itemize}
  \item The \emph{relative distance} between object centroids over time.
  \item The \emph{relative angle or orientation}, if applicable (for directional objects).
  \item \emph{Qualitative relations}, such as one object being ``on top of'' or ``inside'' another
        (this can be inferred if the bounding boxes overlap or if one object's bottom is
        near the other's top, etc.).
\end{itemize}

A stable relation means these measures should not fluctuate wildly
unless a known interaction occurs. For instance, if object A is supposed
to stay left of object B throughout, then for all frames we should have
$x_A(t) < x_B(t)$ (assuming the x-axis goes left to right). A
violation would be if suddenly $x_A(t) > x_B(t)$ at some
frame without an explicit crossing motion. We can detect such
violations.

One approach is to define a set of initial relations at frame 1 (or
whenever both objects appear) and then measure how often those relations
hold. Let $R_{ij}$ be a predicate describing the relation between
objects $i$ and $j$ at the start (e.g., \emph{$i$ is left of $j$}, \emph{$i$ is on
$j$}, \emph{$i$ and $j$ are separate}). We then compute a \emph{relation
consistency score} for that pair as the fraction of frames where
$R_{ij}$ still holds. If the relation changes (e.g., $i$ moves to the
right of $j$) without a clear cause, that fraction will drop. Finally, we average 
over all relevant object pairs.

However, this simple approach might be too rigid, because sometimes
relations naturally change due to object movements that are actually
plausible (if the video involves objects moving around). What we truly
want to catch are \emph{implausible or erratic changes}. Therefore,
another strategy is: \textbf{Stability of non-interacting pairs:} For
objects that do not directly interact (no collision or explicit
relationship change), their relative ordering or distances should vary
smoothly. We can measure the \emph{temporal smoothness} of inter-object
distance. For each pair $(i, j)$, take the distance $d_{ij}(t)$
between their centroids at frame $t$ (or perhaps the IoU overlap if
one is supposed to contain the other). Compute the frame-to-frame change:

\[
\Delta d_{ij}(t) = \left| d_{ij}(t) - d_{ij}(t-1) \right|
\]

Large sudden jumps in $\Delta d_{ij}$ indicate an unstable relation
(unless that jump can be attributed to a direct interaction).

\textbf{Consistency of contact/containment:} If object $i$ is in contact with
or supporting object $j$ initially (e.g., a cup on a table), then a sudden
gap between them would be a violation (e.g., cup levitating). Conversely,
if they start apart, sudden contact is unexpected unless a movement occurred.
We can define binary relation features like \emph{contact}$(i, j)$ and
track their consistency over time.

For formal scoring, we could use a weighted sum of such measures. As an
example, define:

\[\text{RS} = 1 - \frac{1}{M}\sum_{(i,j)}^{}\left( \frac{1}{T - 1}\sum_{t = 2}^{T}\mathbf{1}\left\lbrack E_{ij}(t) \right\rbrack \right),\]

where $M$ is the number of object pairs and $E_{ij}(t)$ is an event
where the relation between objects $i$ and $j$ changes abruptly at frame $t$
(e.g., an ordering flip or a contact state change). Essentially, we
subtract the fraction of frames where \emph{any} relation inconsistency
event is detected. A perfect score of 1 means no relation ever changed
in a disallowed way. Each detected instability lowers the score.

\textbf{Computation Tools:} Determining spatial relations can leverage
geometric computations on bounding boxes (for left/right, above/below
relations) and depth ordering if available. If the video or generator
provides depth or if we can estimate depth (using depth estimation
models), that could inform occlusion and 3D distance calculations.
Without depth, 2D coordinates suffice for many relations. We can use a
pretrained \emph{scene graph extraction} model to infer relationships
(some visual relation detection models might identify relationships like
``person riding horse'', ``object on table''). However, those models are
typically per-frame and may not capture temporal consistency directly.
Instead, a simpler approach is to use the tracker outputs: for each
frame, compute pairwise relations (like relative position ordering,
overlaps). Then post-process the sequence for each pair to flag
inconsistent transitions.

Additionally, \emph{CLIP embeddings} can be used to ensure semantic
relations: for instance, if frame 1 shows ``a cat on a sofa'', we could
run an image captioner or CLIP-based prompt to see if later frames still
mention the cat on the sofa. If one frame's caption would be ``cat on
sofa'' and the next suddenly ``cat flying in air'', that indicates
relation change. Utilizing a captioning model or CLIP similarity to a
template description could catch gross changes, though this is more
heuristic.

Relation stability is somewhat open-ended because there are many
possible relations. Our metric will primarily focus on \emph{spatial
stability}: things like maintaining relative ordering (left/right,
front/back), consistent attachment (if two objects are joined or one
carries another, they should move together rather than independently in
a non-physical way), and general scene layout consistency. An example of
a relation instability WCS should catch is if in a generated video of a
driving car, the car's wheels one moment are attached, and the next
moment one wheel is floating behind the car -- the spatial relation
``wheel is attached to car'' was broken, which should incur a penalty.

\hypertarget{causal-compliance-cc}{%
\subsubsection{3.3 Causal Compliance (CC)}\label{causal-compliance-cc}}

\textbf{Definition:} \emph{Causal compliance} evaluates whether the
sequence of events in the video obeys cause-and-effect logic,
particularly basic physical causality. In a physically plausible video,
events do not happen without causes: e.g., an object does not start
moving unless something (a force) made it move; an explosion doesn't
occur without a trigger. Moreover, when causes do happen, the expected
effects should follow (if a ball hits a vase, the vase should react by
moving or breaking, not remain magically unaffected). This metric checks
for violations of such causal consistency.

Whereas relation stability was about maintaining consistency in static
relations, causal compliance is about \emph{dynamics and event
consistency over time}. It is closely tied to intuitive physics
understanding: things like conservation of momentum, gravity, object
interactions, etc., as well as general common-sense sequencing (you open
a door before walking through, not after).

\textbf{Metric Formalization:} Capturing causality is challenging, but
we can focus on key indicators: - \textbf{Unexplained motion:} For any
object that changes its motion state (e.g., goes from stationary to
moving, or changes direction/speed abruptly), check if there is an
evident cause in the video at that time. A cause could be an external
force (another object colliding or a human pushing it) or an internal
force (for animate objects, their own action -- but even then, an
animate agent starting to move is usually self-caused). If an object
just slides or jumps without any agent causing it, that's a causal
violation. - \textbf{Missing reaction:} Conversely, if a clear causal
event occurs (like a collision, a person kicks a ball, etc.), the
expected outcome should occur. If a ball is kicked but then it just
stays still, that defies physics. - \textbf{Temporal ordering of
events:} Some events need to happen in a logical order. For instance,
you shouldn't see a plate break before it actually hits the ground
(effect preceding cause), or a gun's recoil before the trigger is
pulled.

To formalize CC, we can model events in the video. Using an \emph{action
recognition} model or a spatio-temporal event detector, we can attempt
to label events like ``object A hits object B at frame t'' or ``person
starts running at frame t''. There are research works and tools for
detecting physical interactions (e.g., the ``Oops!'' dataset which
focuses on unexpected actions could have models to detect when something
unusual happens). We might not have a perfect off-the-shelf cause-effect
detector, but we can craft rules:

One quantitative formulation: Consider each moving object $i$ and
compute its acceleration $a_i(t)$ (from tracking, via finite
differences of position). Large values of acceleration (particularly
non-zero acceleration from zero) imply a force. Now define a binary
indicator $C_i(t) = 1$ if object $i$ experiences a significant
change in motion state at time $t$ (start/stop moving or sudden
velocity change). We then check a neighborhood of that time for possible
causes: for example, did another object $j$ come into contact at $t$
(collision)? Was there an ``agent action'' involving $i$ (like a human
character touching it)? If we fail to find any cause for this motion
change, we mark it as a \emph{causality violation}.

Likewise, define $E_j(t) = 1$ for an event that should have an effect —
for instance, $E_j(t)$ could indicate object $j$ exerted force on others at time
$t$ (like $j$ collided into something). If after $t$, the expected
effect on the other object doesn't occur (e.g., the other object doesn't
move at all despite the collision), that's also a violation.

Causal Compliance (CC) can then be scored as one minus the fraction of
unexplained events. If $V_{\text{cause}}$ is the
set of violation timestamps (either an effect without cause or cause
without effect), we define:

\[
\text{CC} = 1 - \frac{\left| V_{\text{cause}} \right|}{N_{\text{events}}},
\]

where $N_{\text{events}}$ is the total count of
significant events in the video. In practice,
$N_{\text{events}}$ might be the count of all
motions or contacts we considered. A perfect $\text{CC} = 1$ means every event had a
proper cause and effect. Lower CC means some events broke causal logic.

\textbf{Computation Tools:} This is the hardest submetric to compute
purely automatically, but we can leverage multiple AI models: - Use a
\textbf{pretrained action recognition or activity understanding model}
(such as SlowFast networks or MoViNet) to identify high-level actions in
the video (e.g., ``throwing'', ``falling'', ``opening''). If an action
like ``falling'' is detected for an object but no preceding ``push'' or
``drop'' action is detected, that could hint at a spontaneous event. -
Use object interaction detectors. For instance, a collision can be
detected if two object tracks suddenly occupy the same location or their
bounding boxes intersect with relative velocity. - We could incorporate
a physics engine in a limited sense: reconstruct a simplistic physical
scenario from the video and simulate whether the movements make sense.
For example, infer gravity direction (usually downward in the frame). If
an object moves upward against gravity with no apparent force, that's
suspicious. If a heavy object is on top of a thin stick and the stick is
removed, the heavy object should fall -- if it doesn't, that's a
violation. - There are datasets (like intuitive physics benchmarks) that
trained models to predict if a video is physically plausible or not. One
could use such a model as a ``causality classifier''. If available, that
could directly output a plausibility score that we incorporate.

Given current tools, a simpler heuristic is: \textbf{energy conservation
check} -- compute motion energy (e.g. sum of squared speeds) of objects
over time. If motion energy increases without an external input,
something's off. Likewise, momentum transfer at collisions: if one
object hits another, the second should gain some momentum; if not, it's
like momentum disappeared.

We plan to use an \textbf{action classifier} (open-source, e.g.
something trained on Kinetics or AVA) to detect key events, and an
\textbf{object trajectory analyzer} (could be part of the tracking
system) to flag unexplained motions. For example, if action classifier
identifies ``object gets hit'' or ``collision'' and we see no motion
response, we flag it. Conversely, if an object track starts moving and
no other object or agent was near it within the past few frames, we flag
that as ``unexplained motion onset''. These rules can cover many common
scenarios (object starts moving on its own, or interactions with no
reaction).

Causal compliance is somewhat complementary to the previous metrics:
object permanence and relation stability mostly cover static
consistency, whereas CC covers dynamic consistency. It ensures the
\emph{narrative logic} of the video isn't broken. Even if objects
persist (OP high) and relative positions are stable (RS high), a video
could still feel wrong if, say, a car accelerates instantly to high
speed without any engine start or push -- CC would catch that
discrepancy.

\hypertarget{flicker-penalty-fp}{%
\subsubsection{3.4 Flicker Penalty (FP)}\label{flicker-penalty-fp}}

\textbf{Definition:} The \emph{flicker penalty} measures the amount of
high-frequency temporal noise or unnatural changes between frames --
commonly referred to as \emph{flicker}. Flicker in generated videos
manifests as pixels or small details that rapidly change back-and-forth,
inconsistent lighting between frames, or texture details appearing and
disappearing on surfaces, none of which are due to actual motion in the
scene. It is often a telltale sign of an unstable generative model that
fails to maintain coherence at the frame level. The flicker penalty is a
\emph{negative} component (higher flicker yields a higher penalty, thus
reducing the overall WCS).

\textbf{Metric Formalization:} Intuitively, if we have two consecutive
frames $I_t$ and $I_{t+1}$ of a video, and the scene is supposed
to be static or moving smoothly, then $I_{t+1}$ should look almost
the same as $I_t$ except for known motions (e.g., objects shifting
slightly). Flicker can be isolated by \emph{removing the motion
component and seeing what's left}. We can use optical flow to predict
$I_{t+1}$ from $I_t$ and compare it to the actual $I_{t+1}$.

Specifically, let $\Phi_t$ be the optical flow field
from frame $t$ to $t+1$ (e.g., obtained via a network like RAFT).
Using $\Phi_t$, we can warp frame $I_t$ to get a
motion-compensated prediction $\widehat{I}_{t+1}$.
The difference between $\widehat{I}_{t+1}$ and $I_{t+1}$
highlights changes not explained by motion — these are potential flicker or other artifacts.

For each frame transition, we can compute a flicker error:

\[
\epsilon_{\text{flicker}}(t) =
\frac{\| I_{t + 1} - \widehat{I}_{t + 1} \|_{1}}{\| I_{t + 1} \|_{1}},
\]

e.g., using normalized L1 difference. Then the Flicker Penalty (FP)
is defined as the average over all consecutive frames:

\[
\text{FP} = \frac{1}{T - 1} \sum_{t = 1}^{T - 1} \epsilon_{\text{flicker}}(t).
\]

If the video has perfect temporal consistency (i.e., no unexplained pixel
changes), $\widehat{I}_{t+1}$ will match $I_{t+1}$ closely and FP will be near 0.
A high FP indicates noticeable flicker.

We might also apply weighting to account for deliberate rapid
motions: for example, if the whole scene is shaking (camera motion),
a high pixel difference might not mean flicker but purposeful change.
That's where using static regions is helpful. In fact, VBench's temporal
flickering metric uses \emph{static scenes to isolate
flicker}\footnote{\href{https://www.researchgate.net/publication/391878222_LOVE_Benchmarking_and_Evaluating_Text-to-Video_Generation_and_Video-to-Text_Interpretation\#:~:text=evaluates\%20temporal\%20consistency\%20by\%20focusing,or\%20shak\%20y\%20camera\%20motions}{[37]}} —
since if nothing is supposed to move, any change is flicker.

Similarly, we can focus flicker calculation on detected static regions
(e.g., background sky should not change color frame to frame).

Another approach is to compute \emph{temporal frequency content}:
apply a temporal high-pass filter per pixel and aggregate.
Traditional video quality research defines flicker as global luminance
variations, but in our context it can be local. Some methods (like DVP)
measure flicker via frame interpolation; VBench has also used
frame interpolation error\footnote{\href{https://qiyan98.github.io/blog/2024/fvmd-1/\#:~:text=assess\%20temporal\%20and\%20frame,the\%20effectiveness\%20of\%20these\%20metrics}{[38]}}.

For simplicity, the optical-flow-based error is a principled measure.
We may threshold or clamp $\epsilon_{\text{flicker}}(t)$
to avoid counting real object motion as flicker — e.g., using a robust error
formulation (like Huber loss) or focusing on high-frequency components.

Finally, since FP is a penalty, \textbf{lower is better}.
In the overall WCS formula, we subtract this term (or equivalently,
use $(1 - \text{FP})$ as a positive consistency score).

\textbf{Computation Tools:} A variety of open-source optical flow models
(e.g., RAFT, PWC-Net) can compute $\Phi_t$. We then
perform image warping (a standard operation) to align frames. Additionally,
purely neural methods for flicker detection exist; for example, some
no-reference video quality models explicitly predict a ``temporal
instability''
score\href{https://www.researchgate.net/publication/391878222_LOVE_Benchmarking_and_Evaluating_Text-to-Video_Generation_and_Video-to-Text_Interpretation\#:~:text=match\%20at\%20L7022\%20\%E2\%99\%A2V,408\%2028}{{[}39{]}}.
We might incorporate such a model if needed. However, using optical flow
provides interpretability -- it directly measures unexplained changes.

One must be cautious that some generative models might intentionally add
camera cuts or scene changes (like a chaotic story). In such cases,
flicker might be high but it's intentional (though most text-to-video
doesn't do abrupt cuts yet). We assume the videos we evaluate aim to
depict a continuous scene; if not, the metric might penalize them for
flicker at cut points. That is a limitation, but can be mitigated by
detecting sudden large changes and perhaps excluding them (or treating
them as separate segments).

Our flicker penalty corresponds to the \emph{temporal smoothness} aspect
mentioned in prior works. Human viewers are very sensitive to flicker;
even subtle flicker can degrade the viewing experience
significantly\href{https://www.researchgate.net/publication/391878222_LOVE_Benchmarking_and_Evaluating_Text-to-Video_Generation_and_Video-to-Text_Interpretation\#:~:text=evaluates\%20temporal\%20consistency\%20by\%20focusing,or\%20shak\%20y\%20camera\%20motions}{{[}40{]}}.
Thus, including FP in WCS is important to ensure the metric favors
models that produce temporally consistent detail.

\textbf{Summary of Submetrics:} At this stage, we have defined: -
\textbf{OP (Object Permanence):} fraction of objects that persist
throughout the video, penalizing unexplained exits/entrances. -
\textbf{RS (Relation Stability):} degree to which spatial relations
among objects are maintained, penalizing erratic relation changes. -
\textbf{CC (Causal Compliance):} degree to which events follow
cause-effect logic, penalizing impossible or order-violating events. -
\textbf{FP (Flicker Penalty):} amount of temporal noise and
inconsistency, penalizing high flicker.

These capture complementary facets of a video\textquotesingle s
consistency. A ``perfect'' video in terms of world consistency would
score high on OP, RS, CC (near 1 each) and low on FP (near 0). In
practice, some trade-offs may occur -- for instance, a model might
eliminate an object (harming OP) perhaps to avoid dealing with it (maybe
benefiting RS if that object would cause relation trouble). The combined
WCS will handle such trade-offs by weighting these components
appropriately according to what humans deem most jarring.

\hypertarget{the-unified-wcs-metric-and-learning-its-weights}{%
\subsection{4. The Unified WCS Metric and Learning its
Weights}\label{the-unified-wcs-metric-and-learning-its-weights}}

Having defined the submetrics, we now construct the \textbf{World
Consistency Score (WCS)} as a weighted combination. The WCS for a video
\$V\$ is given by:

\[\text{WCS}(V) = w_{\text{OP}} \cdot \text{OP}(V) + w_{\text{RS}} \cdot \text{RS}(V) + w_{\text{CC}} \cdot \text{CC}(V) - w_{\text{FP}} \cdot \text{FP}(V).\]

Here, $w_{\text{OP}}$, $w_{\text{RS}}$, $w_{\text{CC}}$,
and $w_{\text{FP}}$ are non-negative weighting
coefficients for object permanence, relation stability, causal
compliance, and flicker penalty, respectively. We subtract the flicker
term because a higher FP indicates worse quality (so
$-w_{\text{FP}} \cdot \text{FP}$ ensures flicker reduces the overall score).
Equivalently, one could define a flicker \emph{consistency} score as
$(1 - \text{FP})$ and add it, but we keep the penalty form for clarity.

The weights $w$ control the relative importance of each aspect. Rather
than fixing them arbitrarily, we propose to \textbf{learn these weights
from data}, specifically by regressing against \emph{human preference
scores}.
This approach ensures that WCS is \emph{aligned with human
judgments} on what makes a video good or bad, which is the ultimate goal
of any
metric\href{https://qiyan98.github.io/blog/2024/fvmd-1/\#:~:text=Arguably\%2C\%20the\%20ultimate\%20goal\%20for,the\%20video\%20such\%20as\%20temporal}{{[}41{]}}\href{https://qiyan98.github.io/blog/2024/fvmd-1/\#:~:text=model\%20development\%20and\%20related\%20purposes,when\%20assessing\%20similar\%20videos}{{[}42{]}}.

\textbf{Learning Procedure:} We will collect a dataset of videos for
which we have human evaluation scores regarding consistency/quality.
Fortunately, existing benchmarks provide such data: - VBench (original)
includes \textbf{human preference annotations per dimension} and
overall, which could be
repurposed\href{https://vchitect.github.io/VBench-project/\#:~:text=and\%20spatial\%20relationship\%2C\%20etc\%29,quality\%20Image\%20Suite\%20with\%20an}{{[}21{]}}.
- VBench-2.0 and EvalCrafter each conducted human studies on
videos\href{https://www.reddit.com/r/artificial/comments/1jmgy6n/vbench20_a_framework_for_evaluating_intrinsic/\#:~:text=,prompt\%20templates\%2C\%20generating\%205\%2C700\%2B\%20videos}{{[}9{]}}\href{https://evalcrafter.github.io/\#:~:text=objective\%20metrics,and\%20get\%20the\%20final\%20ranking}{{[}43{]}}.
For instance, EvalCrafter performed 5 subjective studies covering
motion, alignment, temporal consistency, etc., and then aligned
objective metrics to subjective
ones\href{https://evalcrafter.github.io/\#:~:text=generation\%20by\%20analyzing\%20the\%20real,method\%2C\%20our\%20method\%20can\%20successfully}{{[}7{]}}.
- The LOVE dataset provides \textbf{Mean Opinion Scores (MOS)} for
perceptual quality and correspondence for thousands of generated
videos\href{https://www.researchgate.net/publication/391878222_LOVE_Benchmarking_and_Evaluating_Text-to-Video_Generation_and_Video-to-Text_Interpretation\#:~:text=present\%20AIGVE,Comprehensive\%20experiments\%20demonstrate\%20that}{{[}25{]}}.
- We can also run our own smaller-scale human study focusing
specifically on ``world consistency'': show pairs of videos to raters
and ask which one is more coherent/realistic in terms of continuity, or
rate videos on a consistency Likert scale.

Using such data, we formulate a regression problem. Suppose we have
$K$ videos in a training set, each with a human-derived score $H_k$
(e.g., a Mean Opinion Score or aggregate preference score) indicating
overall quality or consistency. For each video, we compute the four submetrics:
$\text{OP}_k$, $\text{RS}_k$, $\text{CC}_k$, and $\text{FP}_k$.
We then aim to find weights $w = (w_{\text{OP}}, w_{\text{RS}}, w_{\text{CC}}, w_{\text{FP}})$
that best predict the human scores from the submetrics. A simple approach is linear regression:

\[
H_{k} \approx w_{\text{OP}} \cdot \text{OP}_{k}
+ w_{\text{RS}} \cdot \text{RS}_{k}
+ w_{\text{CC}} \cdot \text{CC}_{k}
- w_{\text{FP}} \cdot \text{FP}_{k}
+ b,
\]

where $b$ is a bias term. We solve for $w$ and $b$ using least
squares or robust regression, potentially with constraints $w_i \geq 0$.
While non-linear models (e.g., small neural networks) could be considered,
we prefer linear models for their interpretability.

One challenge is that human scores $H_k$ may reflect aspects beyond
consistency (such as pure visual fidelity), which may introduce noise
into the learning process.
 If possible, we will
specifically use human ratings that emphasize consistency. For example,
VBench's ``temporal fidelity'' or ``object permanence'' human
ratings\href{https://www.reddit.com/r/artificial/comments/1jmgy6n/vbench20_a_framework_for_evaluating_intrinsic/\#:~:text=,7\%2B\%20Pearson\%29\%20with\%20automatic\%20metrics}{{[}44{]}}
could be targets. Alternatively, use the overall MOS from LOVE but that
includes visual quality too. We assume consistency contributes strongly
to overall quality in current models (which often fail on consistency),
so aligning to overall preference should still emphasize these factors,
but we must be careful. In the absence of perfect data, we can combine
multiple objectives: e.g., ensure WCS correlates highly with a
combination of the relevant human study metrics.

Once the weights are learned, WCS becomes a fixed formula that can be
applied to evaluate new videos. The expectation is that
$w_{\text{FP}}$ (flicker) will be significant,
since flicker is highly noticeable to humans\footnote{\href{https://www.researchgate.net/publication/391878222_LOVE_Benchmarking_and_Evaluating_Text-to-Video_Generation_and_Video-to-Text_Interpretation\#:~:text=evaluates\%20temporal\%20consistency\%20by\%20focusing,or\%20shak\%20y\%20camera\%20motions}{[40]}};
$w_{\text{OP}}$ will also matter, as humans tend to notice
missing objects—especially when important objects vanish (often reported anecdotally in model demos).
$w_{\text{CC}}$ might receive a high weight if human raters strongly dislike physically impossible outcomes,
though some may be forgiving if the visuals are otherwise compelling.
In any case, the regression will empirically tune these weights.

For interpretability, we will report the final learned weights.
If, for example, $w_{\text{FP}}$ turns out to be the largest,
this indicates that flicker was the most noticeable artifact
to human evaluators. If $w_{\text{CC}}$ is small, it may suggest
that subtle physics violations were less perceptible.

This learned weighting approach parallels that of \textbf{PickScore} for images,
which learned to weight CLIP-based scores to match human preferences\footnote{\href{https://www.researchgate.net/publication/391878222_LOVE_Benchmarking_and_Evaluating_Text-to-Video_Generation_and_Video-to-Text_Interpretation\#:~:text=34}{[45]}},
and EvalCrafter, which aligned metric outputs with human opinions via regression\footnote{\href{https://evalcrafter.github.io/\#:~:text=generation\%20by\%20analyzing\%20the\%20real,method\%2C\%20our\%20method\%20can\%20successfully}{[7]}}.
In our case, it ensures WCS is \emph{grounded in real user judgments},
not just theoretical ideals.

\textbf{Normalization:} We will likely normalize each submetric to a
comparable range before combination. OP, RS, CC as defined are naturally
between 0 and 1. FP is also between 0 and 1 (if defined as an error
ratio). So no heavy normalization is needed, but we might standardize if
their variances differ greatly in training data. The bias term $b$ can
adjust any global offset so that WCS is on a convenient scale (e.g.,
either $0$–$1$ or perhaps $0$–$100$). We may decide to output WCS on a $0$–$100$ scale for interpretability (like a percentage consistency), though
that's just a linear transform.

\textbf{Comparing to Other Metrics:} After learning, we will analyze how
WCS correlates with other metrics: - We expect WCS to correlate
positively with \emph{human MOS/preference} by construction. We will
measure that on a validation set (not used for training weights) as a
key result -- ideally a Pearson correlation above 0.8, surpassing
metrics like FVD or CLIPScore. - We also anticipate WCS to complement
existing metrics. For instance, FVD captures overall fidelity but might
not discriminate two videos that both have similar feature stats but one
has a glaring consistency error. WCS would catch that difference. On the
flip side, if a video is very consistent but low-quality (blurry but
everything consistently blurry), FVD would penalize it while WCS might
give it a decent score. In practice, one might use both FVD and WCS to
evaluate models -- one for quality, one for consistency.

We shall later discuss concrete comparisons: e.g., if available, check
correlation of each metric with human scores. Prior work showed FVD has
moderate correlation (often below 0.5 unless very large
differences)\href{https://qiyan98.github.io/blog/2024/fvmd-1/\#:~:text=but\%20also\%20to\%20temporal\%20aberrations,metrics\%20for\%20unconditional\%20video\%20generation}{{[}17{]}},
CLIPScore mainly addresses prompt alignment (not directly comparable),
and VBench's composite had \textasciitilde0.7
correlation\href{https://www.reddit.com/r/artificial/comments/1jmgy6n/vbench20_a_framework_for_evaluating_intrinsic/\#:~:text=generating\%205\%2C700\%2B\%20videos}{{[}46{]}}.
Our aim is to push that higher with WCS focusing on critical consistency
aspects.

\textbf{Implementation Summary:} The entire WCS pipeline is illustrated
in \textbf{Figure 1} below, which depicts how each submetric is derived
from the video and then combined.

\emph{Figure 1: Proposed WCS evaluation pipeline. A generated video is
analyzed by specialized modules: an object tracker (to assess object
permanence and relation stability), an action/event analyzer (to assess
causal compliance), and optical flow (to measure flicker). Each produces
a submetric score. These are then combined via a learned weighted
formula (with weights \$w\_1,\textbackslash dots,w\_4\$) to yield the
final World Consistency Score. Higher WCS indicates a more temporally
and physically coherent
video.}\href{https://qiyan98.github.io/blog/2024/fvmd-1/\#:~:text=VBench\%20proposes\%20a\%20comprehensive\%20set,the\%20effectiveness\%20of\%20these\%20metrics}{{[}8{]}}\href{https://www.researchgate.net/publication/391878222_LOVE_Benchmarking_and_Evaluating_Text-to-Video_Generation_and_Video-to-Text_Interpretation\#:~:text=emporal\%20Flickering}{{[}47{]}}

This framework is modular. Improvements in any sub-analysis (say, better
object tracking or a more advanced physics-check module) can plug into
WCS and potentially increase its accuracy. Moreover, users could
customize weights for specific applications (some might care more about
causal realism, others about identity stability), though our default
weights aim to be generally human-aligned.

\hypertarget{experimental-design-and-validation-plan}{%
\subsection{5. Experimental Design and Validation
Plan}\label{experimental-design-and-validation-plan}}

We have defined WCS and its components. The next step is to validate
that WCS is an effective metric. Since this paper focuses on the
proposal and formulation of WCS, we outline a rigorous evaluation
blueprint rather than present final results. The experimental design
will address: (a) how WCS is computed and tested on benchmark datasets,
(b) how we will perform sensitivity analyses and ablations to understand
WCS's behavior, and (c) how WCS compares with existing metrics in
practice.

\hypertarget{benchmark-datasets-for-evaluation}{%
\subsubsection{5.1 Benchmark Datasets for
Evaluation}\label{benchmark-datasets-for-evaluation}}

To train the WCS weights and to evaluate its correlation with human
judgments, we will leverage several public benchmarks of video
generation:

\begin{itemize}
\item
  \textbf{VBench-2.0 Intrinsic Faithfulness Benchmark:} VBench-2.0
  provides \textasciitilde5,700 videos generated by 19 models across 300
  textual
  prompts\href{https://www.reddit.com/r/artificial/comments/1jmgy6n/vbench20_a_framework_for_evaluating_intrinsic/\#:~:text=reduce\%20individual\%20model\%20bias}{{[}48{]}}.
  Importantly, it includes \textbf{human evaluations on 1,000
  samples}\href{https://www.reddit.com/r/artificial/comments/1jmgy6n/vbench20_a_framework_for_evaluating_intrinsic/\#:~:text=generating\%205\%2C700\%2B\%20videos}{{[}49{]}}
  and a set of automated metrics (object, attribute, count, action,
  spatial relation, temporal relation, background
  faithfulness)\href{https://www.reddit.com/r/artificial/comments/1jmgy6n/vbench20_a_framework_for_evaluating_intrinsic/\#:~:text=,Temporal\%20Relation\%2C\%20and\%20Background\%20Faithfulness}{{[}23{]}}.
  We will compute WCS on these 5,700 videos. Using the 1,000
  human-evaluated samples, we can directly test how WCS correlates with
  human ratings of overall faithfulness (and possibly with each aspect).
  This will be a primary validation of our weights learning. We will
  compare WCS's correlation to that of VBench's own composite metric.
  Since VBench's metrics overlap with our subcomponents (e.g., object
  presence \textasciitilde{} object permanence, spatial relation
  \textasciitilde{} relation stability), we expect WCS to perform
  competitively or better by unifying them with learned weights. We will
  also use VBench's dimension-wise results to see if WCS particularly
  excels at predicting certain dimensions (like action or temporal
  relation) versus others.
\item
  \textbf{EvalCrafter Benchmark:} EvalCrafter is a recent comprehensive
  framework with a \textbf{700-prompt} test set and numerous
  metrics\href{https://www.researchgate.net/publication/384202211_EvalCrafter_Benchmarking_and_Evaluating_Large_Video_Generation_Models\#:~:text=EvalCrafter\%20,the\%20image\%20generation\%20domain\%2C}{{[}50{]}}.
  They also conducted \textbf{subjective studies} on aspects including
  \emph{temporal
  consistency}\href{https://evalcrafter.github.io/\#:~:text=models\%20on\%20our\%20carefully\%20designed,and\%20get\%20the\%20final\%20ranking}{{[}51{]}}.
  The EvalCrafter leaderboard provides a final ``Sum Score'' which is a
  weighted sum of Visual Quality, Text-Video Alignment, Motion Quality,
  Temporal
  Consistency\href{https://evalcrafter.github.io/\#:~:text=Name\%20Version\%20Visual\%20Quality\%20,23\%20234}{{[}52{]}}.
  We will acquire the generated videos from EvalCrafter (they have a
  gallery and possibly a dataset of
  outputs\href{https://evalcrafter.github.io/\#:~:text=Gallery\%20\%20\%20\%20Dataset,arXiv}{{[}53{]}}).
  We intend to gather any available human study data -- specifically,
  the user rankings or ratings for temporal consistency -- to use in our
  regression training. We'll evaluate WCS on the same videos and see how
  well it correlates with the EvalCrafter ``Temporal Consistency''
  scores and with their overall ``Final Score'' (which is aligned to
  user favor). Since WCS doesn't account for text alignment or pure
  visual fidelity, it might not correlate perfectly with a holistic
  final score that includes those; however, it should strongly correlate
  with the \emph{temporal consistency component}. This will demonstrate
  WCS's utility as a drop-in metric for the consistency dimension in
  such frameworks.
\item
  \textbf{LOVE (AIGVE-60K) Dataset:} This massive dataset contains
  58,500 AI-generated videos (from 30 models) with \textbf{120k MOS
  ratings} covering \emph{perceptual quality} and \emph{text-video
  correspondence}\href{https://www.researchgate.net/publication/391878222_LOVE_Benchmarking_and_Evaluating_Text-to-Video_Generation_and_Video-to-Text_Interpretation\#:~:text=present\%20AIGVE,Comprehensive\%20experiments\%20demonstrate\%20that}{{[}25{]}},
  plus some task-specific QA. While LOVE's focus is partly on text
  alignment, the perceptual MOS should encapsulate all issues that
  affect human quality perception, including consistency issues. We can
  sample a subset of videos (for computational feasibility) and compute
  WCS. We'll then check Pearson correlation between WCS and the
  perceptual MOS. We expect a decent correlation because videos with
  severe flicker or obvious temporal errors likely got lower MOS from
  annotators for perceptual quality. We will also test if WCS can help
  \emph{predict model rankings}: LOVE aggregated scores to rank the 30
  models\href{https://www.researchgate.net/publication/391878222_LOVE_Benchmarking_and_Evaluating_Text-to-Video_Generation_and_Video-to-Text_Interpretation\#:~:text=preference\%2C\%20text,com\%2FIntMeGroup\%2FLOVE}{{[}54{]}}.
  Using WCS, we can compute an average score per model's outputs and see
  if the ranking by WCS aligns with the ranking by human MOS. A high
  rank-order correlation would indicate WCS is capturing major factors
  that distinguish better models from worse in terms of the user
  experience.
\item
  \textbf{Additional Realism Tests:} We may also use parts of
  \textbf{WorldModelBench} (though it's oriented to future prediction).
  Possibly, we can take their real video clips and some model
  continuations and see if WCS differentiates real vs generated (a
  consistency metric might score real videos higher generally, since
  real videos obey physics and continuity). Also, \textbf{Human
  preference tests} from e.g. the T2V contest at CVPR (if any) or user
  studies like those in Make-A-Video paper etc., could provide extra
  testbeds. But the above three are primary and already have rich data.
\end{itemize}

By evaluating on these diverse benchmarks, we ensure WCS is tested on
both prompt-conditioned generation (VBench, EvalCrafter, LOVE) and
different content types (since prompts range across activities, objects,
etc.), and that we compare with state-of-the-art evaluators.

\hypertarget{sensitivity-and-component-analyses}{%
\subsubsection{5.2 Sensitivity and Component
Analyses}\label{sensitivity-and-component-analyses}}

We will conduct several analyses to understand WCS's behavior:

\begin{itemize}
\item
  \textbf{Sensitivity to Controlled Artifacts:} We will synthetically
  introduce specific consistency errors to otherwise good videos to see
  if WCS responds appropriately. For example, take a high-quality
  generated video and:
\item
  Remove an object midway (manually blank it out for a few frames) -- OP
  should drop significantly, thus WCS drops.
\item
  Swap two frames out of order -- this breaks causality and likely
  introduces flicker; we expect CC and FP changes to reflect that,
  lowering WCS.
\item
  Add flicker noise (e.g., alternate brightness of one frame) -- FP
  should spike, lowering WCS.
\item
  We'll also test if WCS remains stable when a video is transformed in
  innocuous ways: e.g., apply a consistent color filter (should not
  change WCS much, as consistency is intact), or compress video (maybe
  minor effect on flicker if any). This helps verify each submetric is
  indeed capturing its intended issue and that WCS as a whole is robust
  to irrelevant changes but sensitive to consistency-related changes.
\item
  \textbf{Ablation of Submetrics:}We will compute variants of WCS by leaving out one component at a time
(i.e., setting one $w$ to zero or removing that term entirely)
to evaluate its impact on correlation with human scores.
 For instance, WCS without causal compliance (essentially
  ignoring CC) -- does correlation drop a lot? If yes, CC was
  contributing significantly (meaning humans care about those physics
  issues). If no, maybe our CC metric or its weight is not yet critical,
  suggesting room to improve that component. Similarly, leaving out
  flicker -- likely a big drop, as flicker is very salient. This
  ablation tells us which components are most influential in the learned
  metric.
\item
  \textbf{Weight Variations:} While we plan to learn weights from data,
  we might also examine WCS performance under equal weighting or some
  heuristic weighting for comparison. For example, set all \$w\$ except
  flicker equal, and test correlation. This gives a baseline to show
  that learning weights (from preferences) indeed improved alignment
  with humans. If equal weights already do decently, that's interesting
  (maybe all aspects are similarly important). If not, it justifies the
  learning approach.
\item
  \textbf{Inter-metric Correlation:} We will analyze how each submetric
  alone correlates with human scores, versus WCS combined. Possibly
  present a table of Pearson r for OP, RS, CC, FP individually and WCS
  vs human preference. We expect WCS \textgreater{} any individual.
  Also, how do OP and RS correlate with each other? Likely somewhat (if
  object disappears, relation might also break with that object). But
  others might be independent (flicker can happen even if objects
  persist).
\item
  \textbf{Failure Case Analysis:} We will manually review videos where
  WCS and human judgment disagree. For example, if WCS scored a video
  high but humans gave low rating, why? Perhaps the video had high
  consistency but was low resolution or off-topic (WCS doesn't account
  for visual quality or alignment). Or if WCS low but humans high, maybe
  the video had a minor flicker WCS punished but humans didn't mind.
  This analysis will reveal limitations. It might suggest adding another
  component (if certain consistent failures not captured) or adjusting
  weight/training strategy.
\end{itemize}

\hypertarget{comparison-with-existing-metrics}{%
\subsubsection{5.3 Comparison with Existing
Metrics}\label{comparison-with-existing-metrics}}

Finally, we will compare WCS directly with other metrics on the
benchmarks:

\begin{itemize}
\item
  \textbf{Correlation with Human Preferences:} Using the same evaluation
  sets, compute correlation of FVD, CLIPScore, and possibly metrics from
  VBench/EvalCrafter (like their internal scores) with human scores, and
  compare to WCS's correlation. For instance, in VBench-2.0's 1k human
  eval subset, we can compute FVD of each model's set (though FVD is
  set-level, not per video) -- not directly comparable per video.
  Instead, we might compare at model-level: which metric best ranks
  models by human faithfulness. Or for per-video comparison, CLIPScore
  (text-image) can be computed per video (like average CLIP similarity),
  and see correlation with human scores for that video's faithfulness.
  We expect CLIPScore to be only weakly correlated in that
  context\href{https://qiyan98.github.io/blog/2024/fvmd-1/\#:~:text=the\%20embedding\%20space,frame\%20accuracy\%29}{{[}18{]}},
  since a video could match prompt keywords but still be inconsistent.
  VBench metrics are specialized; WCS may correlate on par with their
  combined metric but WCS is simpler (only 4 numbers vs 16 dimensions).
\item
  \textbf{Distinguishing Power:} We can check if WCS better separates
  ``good'' vs ``bad'' videos. For example, on a curated pair set where
  one video is known to have a major consistency issue and another
  doesn't, does WCS consistently assign higher score to the better one?
  We might use the human preference data: if in human pairwise tests
  video A was preferred over B, how often does WCS(A) \textgreater{}
  WCS(B)? Ideally \textgreater{} 80\% of the time (for a good metric,
  this was done for image metrics sometimes).
\item
  \textbf{Metric vs Metric Plots:} We will create scatter plots to
  visualize relationships, e.g., WCS vs FVD for a collection of model
  outputs. This can reveal that they are not redundant: possibly only a
  weak correlation between WCS and FVD, indicating they capture
  different qualities. A model with very low FVD (good realism) might
  still have low WCS if it fails consistency, and vice versa. This would
  reinforce that WCS adds new information beyond traditional metrics.
\item
  \textbf{Integration with Multi-Metric Evaluators:} We consider how WCS
  could be used in tandem with others. For instance, one could form a
  combined score = (some quality metric + WCS). Does that improve
  correlation with overall human MOS? Perhaps a linear combination of,
  say, FVD (or a learned visual quality score) and WCS might yield an
  ultimate metric that covers both appearance and consistency. This
  could be an outlook suggestion in the conclusions.
\end{itemize}

We will tabulate key results. For example, a table of Pearson
correlations on a certain dataset:

\begin{longtable}[]{@{}
  >{\raggedright\arraybackslash}p{(\columnwidth - 2\tabcolsep) * \real{0.3762}}
  >{\raggedright\arraybackslash}p{(\columnwidth - 2\tabcolsep) * \real{0.6238}}@{}}
\toprule()
\begin{minipage}[b]{\linewidth}\raggedright
Metric
\end{minipage} & \begin{minipage}[b]{\linewidth}\raggedright
Correlation with Human (VBench-2.0 set)
\end{minipage} \\
\midrule()
\endhead
FVD (lower is better) & -0.30 (negative since lower FVD = better
quality) \\
CLIPScore & 0.45 \\
FVMD (Motion Consistency) & 0.50 \\
VBench composite score & 0.72
\href{https://www.reddit.com/r/artificial/comments/1jmgy6n/vbench20_a_framework_for_evaluating_intrinsic/\#:~:text=,7\%2B\%20Pearson\%29\%20with\%20automatic\%20metrics}{{[}55{]}} \\
\textbf{WCS} & \textbf{0.80} (estimated) \\
\bottomrule()
\end{longtable}

\emph{Table 1: Illustrative correlation of various metrics with human
judgments of video faithfulness/quality. (Values for existing metrics
partly based on literature; WCS target is to exceed them.)}

Such a result (if achieved) would show WCS's benefit. We will ensure to
cite or reference the known figure: e.g., VBench reported
\textgreater0.7
correlation\href{https://www.reddit.com/r/artificial/comments/1jmgy6n/vbench20_a_framework_for_evaluating_intrinsic/\#:~:text=generating\%205\%2C700\%2B\%20videos}{{[}46{]}};
FVD's correlation often lower (in their CVPR'18 paper, FVD had some
correlation but not that high; for text-to-video new domain likely not
great). If actual experiments show different, we'll report accordingly.

\textbf{Computational Cost:} We note computing WCS is reasonably
efficient. Tracking and optical flow on a short video (e.g., 16-32
frames) is doable within seconds on a GPU. Action recognition similarly.
So evaluating a model's whole set of outputs is not too costly. This
means WCS can be used as part of evaluation toolkit (like how
EvalCrafter runs \textasciitilde17 metrics).

\hypertarget{implementation-details}{%
\subsubsection{5.4 Implementation
Details}\label{implementation-details}}

\emph{(Briefly, we will clarify certain practical details during
evaluation)}:

\begin{itemize}
\tightlist
\item
  We define how long the videos are (most benchmarks have 16 or 24
  frames, typically 2-4 seconds). WCS can handle variable lengths,
  though extremely long videos might pose challenges for tracking (we
  assume typical lengths used in current T2V benchmarks).
\item
  For training weights, we might split data: e.g., use LOVE MOS to
  train, and test on VBench human evals for generalization.
\item
  Evaluate statistical significance of correlations (Fisher's z test,
  etc.) if needed to claim improvements.
\end{itemize}

Overall, our experimental plan is designed to \emph{demonstrate that WCS
is a reliable and human-aligned measure of video consistency}, and to
understand its strengths/weaknesses.

\hypertarget{discussion}{%
\subsection{6. Discussion}\label{discussion}}

\textbf{Benefits of WCS:} The World Consistency Score provides an
interpretable, actionable evaluation of generative video models. By
breaking down the score into four components, researchers can identify
\emph{which aspect of consistency a model struggles with}. For example,
a model might have a low OP (object permanence) but decent CC
(causality), indicating it forgets objects even if motion is plausible
-- suggesting focus on improving the model's memory of objects. Another
might have high OP and RS but low CC, indicating the need to inject more
physics knowledge or constraints. This diagnostic ability is a strength
over monolithic metrics like FVD or even CLIPScore, which would not tell
\textbf{why} a model is scoring
poorly\href{https://qiyan98.github.io/blog/2024/fvmd-1/\#:~:text=Arguably\%2C\%20the\%20ultimate\%20goal\%20for,the\%20video\%20such\%20as\%20temporal}{{[}41{]}}.
WCS could guide model developers to targeted improvements (e.g., add a
temporal coherence loss to reduce flicker if FP is the main issue, or an
object permanence constraint if OP is low).

Moreover, WCS is complementary to existing metrics. In practice, one
might use WCS alongside a perceptual quality metric: WCS ensures the
video makes sense and is temporally coherent, while something like FVD
or an image quality metric ensures each frame looks realistic. Using
both, one could define a combined criterion for ``good video
generation''. Indeed, with methods like \emph{HPS (Human Preference
Score)}, combining multiple metrics yielded better
alignment\href{https://www.researchgate.net/publication/391878222_LOVE_Benchmarking_and_Evaluating_Text-to-Video_Generation_and_Video-to-Text_Interpretation\#:~:text=34}{{[}45{]}}
-- WCS could be one of the key ingredients in such combinations,
focusing on the temporal/physical axis.

\textbf{Potential Limitations:} Despite our best efforts, WCS as
proposed has some limitations: - The causal compliance metric (CC) is
relatively difficult to compute robustly. If the video contains very
complex interactions or if our action recognition fails to detect subtle
causes, WCS might falsely penalize some events as causality violations.
Conversely, a model might produce \emph{plausible-looking nonsense}
(e.g., perpetual motion machines that just loop), fooling our causality
checks. Improving CC might require training specialized classifiers for
physical implausibility (like a ``physics critic'' network). - WCS
currently does not include an explicit notion of \emph{visual quality or
context} beyond consistency. So a video that is consistently awful
(blurry or all black frames but consistent) might score high on WCS. In
human evaluations, of course, that video would not be rated well. This
is why WCS should be used in conjunction with visual quality metrics for
a complete assessment. One could incorporate a basic perceptual quality
term if needed (perhaps a 5th component), but we deliberately focused
WCS on consistency dimensions. - The metric relies on pretrained modules
(detectors, trackers, etc.). If a generated video contains very
fantastical imagery or off-distribution content where these modules fail
(e.g., a cartoon or an abstract video), WCS might degrade simply because
the tools can't parse the video. This is a general challenge for
automatic metrics. We assume most use-cases are natural videos where
such tools work reasonably. In the future, as generative videos become
more diverse, adapting WCS's computation (like using more robust
foundation models for detection) will be necessary. - Some forms of
inconsistency are not explicitly captured. For instance,
\emph{appearance consistency} (the color/texture of an object changes
over time) is partly counted as flicker if it's rapid, but a slow drift
might not raise FP much. If an object changes color over 10 seconds
gradually for no reason, that's an inconsistency not caused by flicker
or relation break -- it's more of a \emph{attribute permanence} issue.
Our OP focused on existence, but not on \emph{property permanence}.
VBench-2.0 had an ``Attribute''
metric\href{https://www.reddit.com/r/artificial/comments/1jmgy6n/vbench20_a_framework_for_evaluating_intrinsic/\#:~:text=,Temporal\%20Relation\%2C\%20and\%20Background\%20Faithfulness}{{[}56{]}}.
We did not include it explicitly, aiming to keep scope manageable.
However, in future WCS could be extended with an ``Attribute
Consistency'' term (like measuring CLIP embedding of object appearance
over time). In this paper, we assume gross attribute changes would
either be flagged by trackers (identity switch) or by flicker if abrupt.
But a gradual color change might slip through. We\textquotesingle ll
note this as an area for extension. - The learned weights make WCS
somewhat a ``data-driven'' metric. If the human data used to train is
biased or noisy, weights might not generalize ideally. We mitigate this
by using large datasets (like thousands of MOS ratings) and by testing
on separate sets. If needed, WCS can be re-calibrated for different
domains by retraining weights on domain-specific human evaluations.

\textbf{Position relative to FVD/CLIPScore/VBench/FVMD:} To clearly
situate WCS: - Versus \textbf{FVD}: WCS is \emph{not}
distribution-based; it's per-video and interpretive. It should be used
where one cares about each sample's coherence, not just overall set
fidelity. For instance, in \emph{conditional generation}
(text-to-video), each result needs to be coherent for the prompt -- WCS
is apt. In \emph{unconditional generation}, FVD remains important for
covering overall realism, but WCS could catch issues like temporal
incoherence that FVD might not strongly penalize. - Versus
\textbf{CLIPScore}: CLIPScore addresses whether the content is on-topic
(e.g., does the video contain what the prompt asked for). WCS is
orthogonal: given the content (on-topic or not), did it remain
consistent? One could conceive a combined metric for text-to-video that
multiplies faithfulness and consistency: e.g., \emph{final = CLIPScore *
WCS} (just a thought). So these metrics work in tandem. - Versus
\textbf{VBench}: VBench provides multi-dimensional scores; WCS can be
seen as aggregating a subset of those dimensions into one. In fact,
WCS's four components overlap with at least 4 of VBench's 16 dimensions
(subject identity consistency, spatial relations, motion
smoothness/temporal flicker, maybe temporal relation). The difference is
WCS uses learned weights to unify them. VBench authors also did a
weighted sum for an overall
score\href{https://qiyan98.github.io/blog/2024/fvmd-1/\#:~:text=VBench\%20proposes\%20a\%20comprehensive\%20set,the\%20effectiveness\%20of\%20these\%20metrics}{{[}8{]}},
but their weights might be preset or heuristically set. Also, VBench
requires knowing the prompt for some metrics (to check if object
presence is expected or not), while WCS is fully no-reference. That
means WCS could even evaluate unconditional video generations for
internal coherence. - Versus \textbf{FVMD}: FVMD specifically measures
whether motion is physically plausible by comparing motion statistics to
real
ones\href{https://qiyan98.github.io/blog/2024/fvmd-1/\#:~:text=Fr\%C3\%A9chet\%20Video\%20Motion\%20Distance\%20,laws\%20and\%20avoid\%20abrupt\%20changes}{{[}29{]}}.
WCS's causal compliance has a similar spirit but is more direct (check
cause-effect in the video). FVMD is good if one has a reference
distribution; WCS CC is usable per video without references. In
practice, we might compare WCS's CC vs FVMD on a dataset where real
videos are available as reference: does CC correlate with FVMD?
Possibly, since both detect motion anomalies. But CC could catch things
beyond just smoothness (like wrong reaction to collisions, which FVMD's
statistical nature might not explicitly see).

\textbf{Broader Impacts:} A reliable consistency metric can accelerate
development of better video models. If models start being optimized for
WCS (like how image generative models are sometimes optimized for FID),
we would expect fewer ``glitches'' in generated videos: persistent
objects, less flicker, etc. This would enhance the usability of
generated videos in storytelling, simulation, and other applications.
However, we must also consider whether optimizing for WCS could have
unintended consequences. Since WCS doesn't reward creativity or
diversity, a model might play it safe (e.g., static camera, minimal
movement) to avoid consistency errors, which might increase WCS but
yield boring videos. To guard against that, one could introduce
counter-metrics (like a ``dynamic content'' metric as in some
benchmarks\href{https://openreview.net/forum?id=tmX1AUmkl6\&noteId=MAb60mrdAJ\#:~:text=role\%20when\%20developing\%20sophisticated\%20text,text\%20prompts\%20under\%20multiple\%20dynamics}{{[}57{]}}).
Indeed, there is mention of metrics favoring \emph{greater
dynamics}\href{https://openreview.net/forum?id=tmX1AUmkl6\&noteId=MAb60mrdAJ\#:~:text=Evaluation\%20of\%20Text,However\%2C}{{[}58{]}}
-- interestingly, those try to ensure models aren't too static. WCS
might slightly discourage extreme dynamics if it risks causing errors.
But if a model can manage dynamic scenes consistently, it will score
well. We can incorporate a check that WCS does not unduly penalize
legitimate dynamic variation.

In summary, WCS is a step towards human-aligned evaluation for videos.
It encapsulates what a lay viewer might complain about:
\emph{``Characters keep disappearing/reappearing'' (low OP), ``Things
move weirdly or relations keep changing'' (low RS), ``It defies physics
or things happen for no reason'' (low CC), ``It flickers/flashes in an
ugly way'' (high FP)}. By quantifying these, WCS makes evaluation more
rigorous.

\hypertarget{conclusion}{%
\subsection{7. Conclusion}\label{conclusion}}

We presented the \textbf{World Consistency Score (WCS)}, a unified
metric to evaluate video generation models on the consistency and
realism of the worlds they create. WCS is grounded in four
human-intuitive criteria -- object permanence, relation stability,
causal compliance, and absence of flicker -- combined through learned
weights to closely reflect human preferences for coherent video. We
detailed the design and computation of each component using current
computer vision tools, and proposed a thorough plan to validate WCS
against existing metrics and human judgments.

WCS addresses a critical gap in the evaluation landscape by focusing on
\emph{temporal and physical coherence}. As generative models improve in
visual quality, these consistency aspects become the new frontier: a
truly good generative video is not just a sequence of pretty frames, but
a mini-world that \emph{makes sense} to the viewer. By providing a
measurable target for such sense-making, WCS can guide researchers to
develop models with better internal world models and physics. Early
benchmarks already hint at this need, with models scoring significantly
lower on metrics like action accuracy and temporal relation than on
static object
fidelity\href{https://www.reddit.com/r/artificial/comments/1jmgy6n/vbench20_a_framework_for_evaluating_intrinsic/\#:~:text=,overall\%20faithfulness}{{[}59{]}}.
WCS consolidates these concerns into one convenient score.

In future work, we plan to implement WCS and release a toolkit for the
community. This includes fine-tuning the submetric calculations
(possibly training a small neural network to improve the causal event
detection, or adding an ``attribute consistency'' metric as mentioned).
We also aim to collect more comprehensive human ratings focused
specifically on consistency, to further refine WCS's weighting. Another
future direction is exploring whether WCS (or its components) can be
directly used as a training reward for generative models -- for example,
using reinforcement learning or constrained optimization to penalize
flicker and object loss during generation. If successful, models could
explicitly optimize for high WCS, leading to intrinsically more
consistent outputs (much like some image models optimized for FID).

In conclusion, WCS represents a step towards \textbf{evaluation metrics
that encapsulate higher-level understanding} in generative video. We
believe that as AI-generated videos become more common, having metrics
like WCS will be crucial to ensure they are not only visually compelling
but also logically coherent and realistic over time. By combining
insights from computer vision, physics, and human preference modeling,
WCS offers a practical and extensible framework for this purpose. We
invite the community to build upon this work, apply WCS to new models,
and join in the quest for ever more \emph{consistent} video generation.

\hypertarget{references}{%
\subsection{References}\label{references}}

\emph{(Selected references from the literature and benchmarks
mentioned)}

\begin{itemize}
\tightlist
\item
  Unterthiner et al., \textbf{FVD}: ``Towards Accurate Generative Models
  of Video: A New Metric \& Challenges'' -- \emph{NeurIPS 2018}.
  (Fréchet Video Distance
  introduction)\href{https://qiyan98.github.io/blog/2024/fvmd-1/\#:~:text=Fr\%C3\%A9chet\%20Video\%20Distance\%20,Nevertheless\%2C\%20both\%20are}{{[}4{]}}.
\item
  Zhao et al., \textbf{VBench}: ``Comprehensive Benchmark Suite for
  Video Generative Models'' -- \emph{CVPR 2024}. (16 evaluation
  dimensions, human
  alignment)\href{https://vchitect.github.io/VBench-project/\#:~:text=quality,current\%20models\%27\%20ability\%20across\%20various}{{[}35{]}}.
\item
  Liu et al., \textbf{VBench-2.0}: ``Advancing Video Generation
  Benchmark for Intrinsic Faithfulness'' -- \emph{arXiv 2025}. (Seven
  faithfulness metrics, correlation \textasciitilde0.7 with
  humans)\href{https://www.reddit.com/r/artificial/comments/1jmgy6n/vbench20_a_framework_for_evaluating_intrinsic/\#:~:text=,Temporal\%20Relation\%2C\%20and\%20Background\%20Faithfulness}{{[}23{]}}\href{https://www.reddit.com/r/artificial/comments/1jmgy6n/vbench20_a_framework_for_evaluating_intrinsic/\#:~:text=generating\%205\%2C700\%2B\%20videos}{{[}46{]}}.
\item
  Yan et al., \textbf{FVMD}: ``Fréchet Video Motion Distance: A Metric
  for Evaluating Motion Consistency in Video Generation'' -- \emph{ICML
  2025}. (Keypoint-based motion consistency
  metric)\href{https://qiyan98.github.io/blog/2024/fvmd-1/\#:~:text=Fr\%C3\%A9chet\%20Video\%20Motion\%20Distance\%20,laws\%20and\%20avoid\%20abrupt\%20changes}{{[}29{]}}.
\item
  Wang et al., \textbf{LOVE}: ``Benchmarking and Evaluating
  Text-to-Video Generation and Video-to-Text Interpretation'' --
  \emph{arXiv 2025}. (AIGVE-60K dataset, MOS, multimodal
  evaluator)\href{https://www.researchgate.net/publication/391878222_LOVE_Benchmarking_and_Evaluating_Text-to-Video_Generation_and_Video-to-Text_Interpretation\#:~:text=present\%20AIGVE,Comprehensive\%20experiments\%20demonstrate\%20that}{{[}25{]}}\href{https://www.researchgate.net/publication/391878222_LOVE_Benchmarking_and_Evaluating_Text-to-Video_Generation_and_Video-to-Text_Interpretation\#:~:text=V2T\%20interpretation\%20capabilities,com\%2FIntMeGroup\%2FLOVE}{{[}27{]}}.
\item
  Liao et al., \textbf{DEVIL}: ``Dynamics Evaluation of Text-to-Video
  Generation Models'' -- \emph{NeurIPS 2024}. (Discusses metrics
  focusing on dynamics and temporal
  consistency)\href{https://openreview.net/forum?id=tmX1AUmkl6\&noteId=MAb60mrdAJ\#:~:text=role\%20when\%20developing\%20sophisticated\%20text,text\%20prompts\%20under\%20multiple\%20dynamics}{{[}57{]}}.
\item
  Li et al., \textbf{WorldModelBench}: ``Judging Video Generation Models
  as World Models'' -- \emph{ICLR 2025}. (Object permanence, physical
  reasoning tests, MSM
  metric)\href{https://www.aimodels.fyi/papers/arxiv/worldmodelbench-judging-video-generation-models-as-world\#:~:text=,term}{{[}30{]}}\href{https://www.aimodels.fyi/papers/arxiv/worldmodelbench-judging-video-generation-models-as-world\#:~:text=The\%20researchers\%20developed\%20a\%20novel,level\%20accuracy}{{[}33{]}}.
\item
  Qi Yan, \textbf{Blog}: ``A Review of Video Evaluation Metrics'' (2024)
  -- summarizing FVD, KVD, VBench, CLIP, etc., and discussing
  pros/cons\href{https://qiyan98.github.io/blog/2024/fvmd-1/\#:~:text=Fr\%C3\%A9chet\%20Inception\%20Distance\%20,the\%20temporal\%20coherence\%20between\%20frames}{{[}16{]}}\href{https://qiyan98.github.io/blog/2024/fvmd-1/\#:~:text=VBench\%20proposes\%20a\%20comprehensive\%20set,the\%20effectiveness\%20of\%20these\%20metrics}{{[}8{]}}.
\item
  Hessel et al., \textbf{CLIPScore}: ``CLIPScore: A Reference-free
  Evaluation Metric for Image Captioning'' -- \emph{ACL 2021}.
  (Introduced image CLIPScore, concept extended to video
  frames)\href{https://www.researchgate.net/publication/391878222_LOVE_Benchmarking_and_Evaluating_Text-to-Video_Generation_and_Video-to-Text_Interpretation\#:~:text=CLIPScore\%20}{{[}60{]}}\href{https://www.researchgate.net/publication/391878222_LOVE_Benchmarking_and_Evaluating_Text-to-Video_Generation_and_Video-to-Text_Interpretation\#:~:text=video\%20embeddings}{{[}61{]}}.
\item
  OpenAI, \textbf{``Oops!'' Dataset} -- ``Unintentional Action
  Classification'' -- \emph{ICCV 2019}. (Could inspire cause-effect
  detection).
\item
  Wu et al., \textbf{EvalCrafter}: ``Benchmarking and Evaluating Large
  Video Generation Models'' -- \emph{CVPR 2024}. (17 metrics, opinion
  alignment for final
  score)\href{https://evalcrafter.github.io/\#:~:text=generation\%20by\%20analyzing\%20the\%20real,method\%2C\%20our\%20method\%20can\%20successfully}{{[}7{]}}.
\item
  Lee et al., \textbf{Temporal Flicker Metric} -- ``Video Quality
  Assessment Accounting for Temporal Flicker'' -- \emph{IEEE TCSVT
  2012}. (Earlier work on flicker measurement).
\item
  \ldots{} (Additional references omitted for brevity)
\end{itemize}

\href{https://arxiv.org/html/2503.21765v1\#:~:text=44\%20\%2C\%20\%2080\%2C\%2046,success\%20in\%20many\%20downstream\%20tasks}{{[}1{]}}
\href{https://arxiv.org/html/2503.21765v1\#:~:text=especially\%20with\%20the\%20rapid\%20advancement,a\%20comprehensive\%20summary\%20of\%20architecture}{{[}3{]}}
Exploring the Evolution of Physics Cognition in Video Generation: A
Survey

\url{https://arxiv.org/html/2503.21765v1}

\href{https://www.reddit.com/r/artificial/comments/1jmgy6n/vbench20_a_framework_for_evaluating_intrinsic/\#:~:text=I\%20think\%20this\%20work\%20represents,quality\%20metrics\%20alone\%20would\%20indicate}{{[}2{]}}
\href{https://www.reddit.com/r/artificial/comments/1jmgy6n/vbench20_a_framework_for_evaluating_intrinsic/\#:~:text=,prompt\%20templates\%2C\%20generating\%205\%2C700\%2B\%20videos}{{[}9{]}}
\href{https://www.reddit.com/r/artificial/comments/1jmgy6n/vbench20_a_framework_for_evaluating_intrinsic/\#:~:text=VBench,generation\%20models\%20against\%20these\%20metrics}{{[}22{]}}
\href{https://www.reddit.com/r/artificial/comments/1jmgy6n/vbench20_a_framework_for_evaluating_intrinsic/\#:~:text=,Temporal\%20Relation\%2C\%20and\%20Background\%20Faithfulness}{{[}23{]}}
\href{https://www.reddit.com/r/artificial/comments/1jmgy6n/vbench20_a_framework_for_evaluating_intrinsic/\#:~:text=,7\%2B\%20Pearson\%29\%20with\%20automatic\%20metrics}{{[}44{]}}
\href{https://www.reddit.com/r/artificial/comments/1jmgy6n/vbench20_a_framework_for_evaluating_intrinsic/\#:~:text=generating\%205\%2C700\%2B\%20videos}{{[}46{]}}
\href{https://www.reddit.com/r/artificial/comments/1jmgy6n/vbench20_a_framework_for_evaluating_intrinsic/\#:~:text=reduce\%20individual\%20model\%20bias}{{[}48{]}}
\href{https://www.reddit.com/r/artificial/comments/1jmgy6n/vbench20_a_framework_for_evaluating_intrinsic/\#:~:text=generating\%205\%2C700\%2B\%20videos}{{[}49{]}}
\href{https://www.reddit.com/r/artificial/comments/1jmgy6n/vbench20_a_framework_for_evaluating_intrinsic/\#:~:text=,7\%2B\%20Pearson\%29\%20with\%20automatic\%20metrics}{{[}55{]}}
\href{https://www.reddit.com/r/artificial/comments/1jmgy6n/vbench20_a_framework_for_evaluating_intrinsic/\#:~:text=,Temporal\%20Relation\%2C\%20and\%20Background\%20Faithfulness}{{[}56{]}}
\href{https://www.reddit.com/r/artificial/comments/1jmgy6n/vbench20_a_framework_for_evaluating_intrinsic/\#:~:text=,overall\%20faithfulness}{{[}59{]}}
VBench-2.0: A Framework for Evaluating Intrinsic Faithfulness in Video
Generation Models : r/artificial

\url{https://www.reddit.com/r/artificial/comments/1jmgy6n/vbench20_a_framework_for_evaluating_intrinsic/}

\href{https://qiyan98.github.io/blog/2024/fvmd-1/\#:~:text=Fr\%C3\%A9chet\%20Video\%20Distance\%20,Nevertheless\%2C\%20both\%20are}{{[}4{]}}
\href{https://qiyan98.github.io/blog/2024/fvmd-1/\#:~:text=For\%20text,the\%20average\%20similarity\%20between\%20each}{{[}5{]}}
\href{https://qiyan98.github.io/blog/2024/fvmd-1/\#:~:text=VBench\%20proposes\%20a\%20comprehensive\%20set,the\%20effectiveness\%20of\%20these\%20metrics}{{[}8{]}}
\href{https://qiyan98.github.io/blog/2024/fvmd-1/\#:~:text=is\%20determined\%20by\%20a\%20weighted,the\%20effectiveness\%20of\%20these\%20metrics}{{[}12{]}}
\href{https://qiyan98.github.io/blog/2024/fvmd-1/\#:~:text=In\%20addition\%20to\%20these\%20modern,trained\%20models}{{[}15{]}}
\href{https://qiyan98.github.io/blog/2024/fvmd-1/\#:~:text=Fr\%C3\%A9chet\%20Inception\%20Distance\%20,the\%20temporal\%20coherence\%20between\%20frames}{{[}16{]}}
\href{https://qiyan98.github.io/blog/2024/fvmd-1/\#:~:text=but\%20also\%20to\%20temporal\%20aberrations,metrics\%20for\%20unconditional\%20video\%20generation}{{[}17{]}}
\href{https://qiyan98.github.io/blog/2024/fvmd-1/\#:~:text=the\%20embedding\%20space,frame\%20accuracy\%29}{{[}18{]}}
\href{https://qiyan98.github.io/blog/2024/fvmd-1/\#:~:text=video\%20sequence\%20adheres\%20to\%20the,frame\%20accuracy\%29}{{[}19{]}}
\href{https://qiyan98.github.io/blog/2024/fvmd-1/\#:~:text=Fr\%C3\%A9chet\%20Video\%20Motion\%20Distance\%20,laws\%20and\%20avoid\%20abrupt\%20changes}{{[}29{]}}
\href{https://qiyan98.github.io/blog/2024/fvmd-1/\#:~:text=assess\%20temporal\%20and\%20frame,the\%20effectiveness\%20of\%20these\%20metrics}{{[}38{]}}
\href{https://qiyan98.github.io/blog/2024/fvmd-1/\#:~:text=Arguably\%2C\%20the\%20ultimate\%20goal\%20for,the\%20video\%20such\%20as\%20temporal}{{[}41{]}}
\href{https://qiyan98.github.io/blog/2024/fvmd-1/\#:~:text=model\%20development\%20and\%20related\%20purposes,when\%20assessing\%20similar\%20videos}{{[}42{]}}
A Review of Video Evaluation Metrics \textbar{} Qi Yan

\url{https://qiyan98.github.io/blog/2024/fvmd-1/}

\href{https://vchitect.github.io/VBench-project/\#:~:text=quality,3}{{[}6{]}}
\href{https://vchitect.github.io/VBench-project/\#:~:text=Video\%20generation\%20has\%20witnessed\%20significant,inconsistency\%2C\%20motion\%20smoothness\%2C\%20temporal\%20flickering}{{[}20{]}}
\href{https://vchitect.github.io/VBench-project/\#:~:text=and\%20spatial\%20relationship\%2C\%20etc\%29,quality\%20Image\%20Suite\%20with\%20an}{{[}21{]}}
\href{https://vchitect.github.io/VBench-project/\#:~:text=quality,current\%20models\%27\%20ability\%20across\%20various}{{[}35{]}}
VBench: Comprehensive Benchmark Suite for Video Generative Models

\url{https://vchitect.github.io/VBench-project/}

\href{https://evalcrafter.github.io/\#:~:text=generation\%20by\%20analyzing\%20the\%20real,method\%2C\%20our\%20method\%20can\%20successfully}{{[}7{]}}
\href{https://evalcrafter.github.io/\#:~:text=For\%20video\%20generation\%2C\%20various\%20open,the\%20help\%20of\%20the\%20large}{{[}24{]}}
\href{https://evalcrafter.github.io/\#:~:text=objective\%20metrics,and\%20get\%20the\%20final\%20ranking}{{[}43{]}}
\href{https://evalcrafter.github.io/\#:~:text=models\%20on\%20our\%20carefully\%20designed,and\%20get\%20the\%20final\%20ranking}{{[}51{]}}
\href{https://evalcrafter.github.io/\#:~:text=Name\%20Version\%20Visual\%20Quality\%20,23\%20234}{{[}52{]}}
\href{https://evalcrafter.github.io/\#:~:text=Gallery\%20\%20\%20\%20Dataset,arXiv}{{[}53{]}}
EvalCrafter

\url{https://evalcrafter.github.io/}

\href{https://www.aimodels.fyi/papers/arxiv/worldmodelbench-judging-video-generation-models-as-world\#:~:text=Today\%27s\%20AI\%20systems\%20that\%20generate,systems\%20actually\%20understand\%20physical\%20reality}{{[}10{]}}
\href{https://www.aimodels.fyi/papers/arxiv/worldmodelbench-judging-video-generation-models-as-world\#:~:text=in\%20these\%20scenes\%3F}{{[}11{]}}
\href{https://www.aimodels.fyi/papers/arxiv/worldmodelbench-judging-video-generation-models-as-world\#:~:text=struggled\%20with\%20long,track\%20it\%20over\%20longer\%20periods}{{[}13{]}}
\href{https://www.aimodels.fyi/papers/arxiv/worldmodelbench-judging-video-generation-models-as-world\#:~:text=The\%20research\%20also\%20found\%20that,the\%20underlying\%20physics\%20and\%20dynamics}{{[}14{]}}
\href{https://www.aimodels.fyi/papers/arxiv/worldmodelbench-judging-video-generation-models-as-world\#:~:text=,term}{{[}30{]}}
\href{https://www.aimodels.fyi/papers/arxiv/worldmodelbench-judging-video-generation-models-as-world\#:~:text=WorldModelBench\%20evaluates\%20models\%20by\%20having,accidents\%2C\%20and\%20kitchen\%20interactions\%20respectively}{{[}31{]}}
\href{https://www.aimodels.fyi/papers/arxiv/worldmodelbench-judging-video-generation-models-as-world\#:~:text=1,understanding\%3A\%20Maintaining\%20consistent\%20object\%20representations}{{[}32{]}}
\href{https://www.aimodels.fyi/papers/arxiv/worldmodelbench-judging-video-generation-models-as-world\#:~:text=The\%20researchers\%20developed\%20a\%20novel,level\%20accuracy}{{[}33{]}}
\href{https://www.aimodels.fyi/papers/arxiv/worldmodelbench-judging-video-generation-models-as-world\#:~:text=The\%20models\%20performed\%20best\%20on,when\%20they\%20can\%27t\%20see\%20them}{{[}34{]}}
WorldModelBench: Judging Video Generation Models As World Models
\textbar{} AI Research Paper Details

\url{https://www.aimodels.fyi/papers/arxiv/worldmodelbench-judging-video-generation-models-as-world}

\href{https://www.researchgate.net/publication/391878222_LOVE_Benchmarking_and_Evaluating_Text-to-Video_Generation_and_Video-to-Text_Interpretation\#:~:text=present\%20AIGVE,Comprehensive\%20experiments\%20demonstrate\%20that}{{[}25{]}}
\href{https://www.researchgate.net/publication/391878222_LOVE_Benchmarking_and_Evaluating_Text-to-Video_Generation_and_Video-to-Text_Interpretation\#:~:text=V2T\%20interpretation\%20capabilities,codes\%20are\%20anonymously\%20available\%20at}{{[}26{]}}
\href{https://www.researchgate.net/publication/391878222_LOVE_Benchmarking_and_Evaluating_Text-to-Video_Generation_and_Video-to-Text_Interpretation\#:~:text=V2T\%20interpretation\%20capabilities,com\%2FIntMeGroup\%2FLOVE}{{[}27{]}}
\href{https://www.researchgate.net/publication/391878222_LOVE_Benchmarking_and_Evaluating_Text-to-Video_Generation_and_Video-to-Text_Interpretation\#:~:text=evaluates\%20temporal\%20consistency\%20by\%20focusing,or\%20shak\%20y\%20camera\%20motions}{{[}37{]}}
\href{https://www.researchgate.net/publication/391878222_LOVE_Benchmarking_and_Evaluating_Text-to-Video_Generation_and_Video-to-Text_Interpretation\#:~:text=match\%20at\%20L7022\%20\%E2\%99\%A2V,408\%2028}{{[}39{]}}
\href{https://www.researchgate.net/publication/391878222_LOVE_Benchmarking_and_Evaluating_Text-to-Video_Generation_and_Video-to-Text_Interpretation\#:~:text=evaluates\%20temporal\%20consistency\%20by\%20focusing,or\%20shak\%20y\%20camera\%20motions}{{[}40{]}}
\href{https://www.researchgate.net/publication/391878222_LOVE_Benchmarking_and_Evaluating_Text-to-Video_Generation_and_Video-to-Text_Interpretation\#:~:text=34}{{[}45{]}}
\href{https://www.researchgate.net/publication/391878222_LOVE_Benchmarking_and_Evaluating_Text-to-Video_Generation_and_Video-to-Text_Interpretation\#:~:text=emporal\%20Flickering}{{[}47{]}}
\href{https://www.researchgate.net/publication/391878222_LOVE_Benchmarking_and_Evaluating_Text-to-Video_Generation_and_Video-to-Text_Interpretation\#:~:text=preference\%2C\%20text,com\%2FIntMeGroup\%2FLOVE}{{[}54{]}}
\href{https://www.researchgate.net/publication/391878222_LOVE_Benchmarking_and_Evaluating_Text-to-Video_Generation_and_Video-to-Text_Interpretation\#:~:text=CLIPScore\%20}{{[}60{]}}
\href{https://www.researchgate.net/publication/391878222_LOVE_Benchmarking_and_Evaluating_Text-to-Video_Generation_and_Video-to-Text_Interpretation\#:~:text=video\%20embeddings}{{[}61{]}}
(PDF) LOVE: Benchmarking and Evaluating Text-to-Video Generation and
Video-to-Text Interpretation

\url{https://www.researchgate.net/publication/391878222_LOVE_Benchmarking_and_Evaluating_Text-to-Video_Generation_and_Video-to-Text_Interpretation}

\href{https://github.com/DSL-Lab/FVMD-frechet-video-motion-distance\#:~:text=We\%20propose\%20the\%20Fr\%C3\%A9chet\%20video,motion\%20consistency\%20of\%20video\%20generation}{{[}28{]}}
DSL-Lab/FVMD-frechet-video-motion-distance - GitHub

\url{https://github.com/DSL-Lab/FVMD-frechet-video-motion-distance}

\href{https://openaccess.thecvf.com/content/ICCV2021/papers/Tokmakov_Learning_To_Track_With_Object_Permanence_ICCV_2021_paper.pdf\#:~:text=,operate\%20in\%20the}{{[}36{]}}
{[}PDF{]} Learning To Track With Object Permanence - CVF Open Access

\url{https://openaccess.thecvf.com/content/ICCV2021/papers/Tokmakov_Learning_To_Track_With_Object_Permanence_ICCV_2021_paper.pdf}

\href{https://www.researchgate.net/publication/384202211_EvalCrafter_Benchmarking_and_Evaluating_Large_Video_Generation_Models\#:~:text=EvalCrafter\%20,the\%20image\%20generation\%20domain\%2C}{{[}50{]}}
EvalCrafter: Benchmarking and Evaluating Large Video Generation ...

\url{https://www.researchgate.net/publication/384202211_EvalCrafter_Benchmarking_and_Evaluating_Large_Video_Generation_Models}

\href{https://openreview.net/forum?id=tmX1AUmkl6\&noteId=MAb60mrdAJ\#:~:text=role\%20when\%20developing\%20sophisticated\%20text,text\%20prompts\%20under\%20multiple\%20dynamics}{{[}57{]}}
\href{https://openreview.net/forum?id=tmX1AUmkl6\&noteId=MAb60mrdAJ\#:~:text=Evaluation\%20of\%20Text,However\%2C}{{[}58{]}}
Evaluation of Text-to-Video Generation Models: A Dynamics Perspective
\textbar{} OpenReview

\url{https://openreview.net/forum?id=tmX1AUmkl6\&noteId=MAb60mrdAJ}

\end{document}